\documentclass[journal]{IEEEtran}

\usepackage{cite}
\usepackage{graphicx}
\usepackage[breaklinks=false, allcolors=black, colorlinks=true]{hyperref}
\usepackage{amsfonts}
\usepackage{amsmath} % Needs to be defined before amsthm package
\usepackage{amssymb}
\usepackage{amsthm}	% Needs to be defined after amsmath package
\usepackage{enumerate}
\usepackage{mathrsfs}
\usepackage{comment}

\usepackage{dsfont}
\usepackage{epstopdf}
\usepackage{subfig}
\usepackage{dashrule}
\usepackage{soul} 
\newcommand{\wholesale}{w}

\theoremstyle{definition} % plain, {definition}, remark (bold title, regular text)
\newtheorem{theore}{Theorem}
\newtheorem{corollar}{Corollary}
\newtheorem{lemm}{Lemma}
\newtheorem{definitio}{Definition}
\newtheorem{proble}{Problem}
\newtheorem{assumptio}{Assumption}
\newtheorem{propert}{Property}
\newtheorem{propositio}{Proposition}
\newtheorem{conjectur}{Conjecture}
\newtheorem{exampl}{Example}
\newtheorem{rul}{Rule}

\newtheorem{remar}{Remark}
\newtheorem{fac}{Fact}

\def\nn{\nonumber}
\def\beq{\begin{equation}}
\def\eeq{\end{equation}}
\def\beqs{\begin{equation*}}
\def\eeqs{\end{equation*}}
\def\balign{\begin{align}}
\def\ealign{\end{align}}
\def\bea{\begin{eqnarray}}
\def\eea{\end{eqnarray}}
\def\ba{\begin{array}}
\def\ea{\end{array}}
\def\bcenter{\begin{center}}
\def\ecenter{\end{center}}
\def\bitem{\begin{itemize}}
\def\eitem{\end{itemize}}
\def\benum{\begin{enumerate}}
\def\eenum{\end{enumerate}}
\def\bdoc{\begin{document}}
\def\edoc{\end{document}}
\def\defeq{{\stackrel{\Delta}{=}}}
\def\nn{\nonumber}

\def\eg{{e.g., \/}}
\def\etal{{\it et al. \/}}
\def\viz{{\iet viz.,\ \/}}
\def\ie{{i.e.,\ \/}}
\def\etc{{\it etc.,\ \/}}
\def\cf{{cf.,\ \/}}

\usepackage[usenames,dvipsnames,svgnames,table]{xcolor}

\def\black{\textcolor{black}}
\def\yellow{\textcolor{yellow}}
\def\green{\textcolor{green}}
\def\tcbg{\textcolor{bgrd}}
\def\magenta{\textcolor{magenta}}
\def\white{\textcolor{white}}
\def\gray{\textcolor{gray}}

\newcommand\Quote[1]{\lq\textsl{#1}\rq}
\newcommand\fr[2]{{\textstyle\frac{#1}{#2}}}

\newcommand{\Ambb}{\mathbb{A}}
\newcommand{\Bmbb}{\mathbb{B}}
\newcommand{\Cmbb}{\mathbb{C}}
\newcommand{\Dmbb}{\mathbb{D}}
\newcommand{\Embb}{\mathbb{E}}
\newcommand{\Fmbb}{\mathbb{F}}
\newcommand{\Gmbb}{\mathbb{G}}
\newcommand{\Hmbb}{\mathbb{H}}
\newcommand{\Imbb}{\mathbb{I}}
\newcommand{\Jmbb}{\mathbb{J}}
\newcommand{\Kmbb}{\mathbb{K}}
\newcommand{\Lmbb}{\mathbb{L}}
\newcommand{\Mmbb}{\mathbb{M}}
\newcommand{\Nmbb}{\mathbb{N}}
\newcommand{\Ombb}{\mathbb{O}}
\newcommand{\Pmbb}{\mathbb{P}}
\newcommand{\Qmbb}{\mathbb{Q}}
\newcommand{\Rmbb}{\mathbb{R}}
\newcommand{\Smbb}{\mathbb{S}}
\newcommand{\Tmbb}{\mathbb{T}}
\newcommand{\Umbb}{\mathbb{U}}
\newcommand{\Vmbb}{\mathbb{V}}
\newcommand{\Wmbb}{\mathbb{W}}
\newcommand{\Xmbb}{\mathbb{X}}
\newcommand{\Ymbb}{\mathbb{Y}}
\newcommand{\Zmbb}{\mathbb{Z}}

\newcommand{\Amsc}{\mathcal{A}}
\newcommand{\Bmsc}{\mathcal{B}}
\newcommand{\Cmsc}{\mathcal{C}}
\newcommand{\Dmsc}{\mathcal{D}}
\newcommand{\Emsc}{\mathcal{E}}
\newcommand{\Fmsc}{\mathcal{F}}
\newcommand{\Gmsc}{\mathcal{G}}
\newcommand{\Hmsc}{\mathcal{H}}
\newcommand{\Imsc}{\mathcal{I}}
\newcommand{\Jmsc}{\mathcal{J}}
\newcommand{\Kmsc}{\mathcal{K}}
\newcommand{\Lmsc}{\mathcal{L}}
\newcommand{\Mmsc}{\mathcal{M}}
\newcommand{\Nmsc}{\mathcal{N}}
\newcommand{\Omsc}{\mathcal{O}}
\newcommand{\Pmsc}{\mathcal{P}}
\newcommand{\Qmsc}{\mathcal{Q}}
\newcommand{\Rmsc}{\mathcal{R}}
\newcommand{\Smsc}{\mathcal{S}}
\newcommand{\Tmsc}{\mathcal{T}}
\newcommand{\Umsc}{\mathcal{U}}
\newcommand{\Vmsc}{\mathcal{V}}
\newcommand{\Wmsc}{\mathcal{W}}
\newcommand{\Xmsc}{\mathcal{X}}
\newcommand{\Ymsc}{\mathcal{Y}}
\newcommand{\Zmsc}{\mathcal{Z}}

\newcommand{\Ascr}{\mathscr{A}}
\newcommand{\Bscr}{\mathscr{B}}
\newcommand{\Cscr}{\mathscr{C}}
\newcommand{\Dscr}{\mathscr{D}}
\newcommand{\Escr}{\mathscr{E}}
\newcommand{\Fscr}{\mathscr{F}}
\newcommand{\Gscr}{\mathscr{G}}
\newcommand{\Hscr}{\mathscr{H}}
\newcommand{\Iscr}{\mathscr{I}}
\newcommand{\Jscr}{\mathscr{J}}
\newcommand{\Kscr}{\mathscr{K}}
\newcommand{\Lscr}{\mathscr{L}}
\newcommand{\Mscr}{\mathscr{M}}
\newcommand{\Nscr}{\mathscr{N}}
\newcommand{\Oscr}{\mathscr{O}}
\newcommand{\Pscr}{\mathscr{P}}
\newcommand{\Qscr}{\mathscr{Q}}
\newcommand{\Rscr}{\mathscr{R}}
\newcommand{\Sscr}{\mathscr{S}}
\newcommand{\Tscr}{\mathscr{T}}
\newcommand{\Uscr}{\mathscr{U}}
\newcommand{\Vscr}{\mathscr{V}}
\newcommand{\Wscr}{\mathscr{W}}
\newcommand{\Xscr}{\mathscr{X}}
\newcommand{\Yscr}{\mathscr{Y}}
\newcommand{\Zscr}{\mathscr{Z}}

\def\Atiny{\mbox{\tiny{A}}}
\def\Btiny{\mbox{\tiny{B}}}
\def\Ctiny{\mbox{\tiny{C}}}
\def\Dtiny{\mbox{\tiny{D}}}
\def\Etiny{\mbox{\tiny{E}}}
\def\Ctiny{\mbox{\tiny{C}}}
\def\Ftiny{\mbox{\tiny{F}}}
\def\Gtiny{\mbox{\tiny{G}}}
\def\Htiny{\mbox{\tiny{H}}}
\def\Itiny{\mbox{\tiny{I}}}
\def\Jtiny{\mbox{\tiny{J}}}
\def\Ktiny{\mbox{\tiny{K}}}
\def\Ltiny{\mbox{\tiny{L}}}
\def\Mtiny{\mbox{\tiny{M}}}
\def\Ntiny{\mbox{\tiny{N}}}
\def\Otiny{\mbox{\tiny{O}}}
\def\Ptiny{\mbox{\tiny{P}}}
\def\Qtiny{\mbox{\tiny{Q}}}
\def\Rtiny{\mbox{\tiny{R}}}
\def\Stiny{\mbox{\tiny{S}}}
\def\Ttiny{\mbox{\tiny{T}}}
\def\Utiny{\mbox{\tiny{U}}}
\def\Vtiny{\mbox{\tiny{V}}}
\def\Wtiny{\mbox{\tiny{W}}}
\def\Xtiny{\mbox{\tiny{X}}}
\def\Ytiny{\mbox{\tiny{Y}}}
\def\Ztiny{\mbox{\tiny{Z}}}

\def\alphabf{\hbox{\boldmath$\alpha$\unboldmath}}
\def\betabf{\hbox{\boldmath$\beta$\unboldmath}}
\def\gammabf{\hbox{\boldmath$\gamma$\unboldmath}}
\def\deltabf{\hbox{\boldmath$\delta$\unboldmath}}
\def\epsilonbf{\hbox{\boldmath$\epsilon$\unboldmath}}
\def\zetabf{\hbox{\boldmath$\zeta$\unboldmath}}
\def\etabf{\hbox{\boldmath$\eta$\unboldmath}}
\def\iotabf{\hbox{\boldmath$\iota$\unboldmath}}
\def\kappabf{\hbox{\boldmath$\kappa$\unboldmath}}
\def\lambdabf{\hbox{\boldmath$\lambda$\unboldmath}}
\def\mubf{\hbox{\boldmath$\mu$\unboldmath}}
\def\nubf{\hbox{\boldmath$\nu$\unboldmath}}
\def\xibf{\hbox{\boldmath$\xi$\unboldmath}}
\def\pibf{\hbox{\boldmath$\pi$\unboldmath}}
\def\rhobf{\hbox{\boldmath$\rho$\unboldmath}}
\def\sigmabf{\hbox{\boldmath$\sigma$\unboldmath}}
\def\taubf{\hbox{\boldmath$\tau$\unboldmath}}
\def\upsilonbf{\hbox{\boldmath$\upsilon$\unboldmath}}
\def\phibf{\hbox{\boldmath$\phi$\unboldmath}}
\def\chibf{\hbox{\boldmath$\chi$\unboldmath}}
\def\psibf{\hbox{\boldmath$\psi$\unboldmath}}
\def\omegabf{\hbox{\boldmath$\omega$\unboldmath}}
\def\Sigmabf{\hbox{$\bf \Sigma$}}
\def\Upsilonbf{\hbox{$\bf \Upsilon$}}
\def\Omegabf{\hbox{$\bf \Omega$}}
\def\Deltabf{\hbox{$\bf \Delta$}}
\def\Gammabf{\hbox{$\bf \Gamma$}}
\def\Thetabf{\hbox{$\bf \Theta$}}
\def\Lambdabf{\mbox{$ \bf \Lambda $}}
\def\Xibf{\hbox{\bf$\Xi$}}
\def\Pibf{{\bf \Pi}}
\def\thetabf{{\mbox{\boldmath$\theta$\unboldmath}}}
\def\Upsilonbf{\hbox{\boldmath$\Upsilon$\unboldmath}}
\newcommand{\Phibf}{\mbox{${\bf \Phi}$}}
\newcommand{\Psibf}{\mbox{${\bf \Psi}$}}

\def\abf{{\bf a}}
\def\bbf{{\bf b}}
\def\cbf{{\bf c}}
\def\dbf{{\bf d}}
\def\ebf{{\bf e}}
\def\fbf{{\bf f}}
\def\gbf{{\bf g}}
\def\hbf{{\bf h}}
\def\ibf{{\bf i}}
\def\jbf{{\bf j}}
\def\kbf{{\bf k}}
\def\lbf{{\bf l}}
\def\mbf{{\bf m}}
\def\nbf{{\bf n}}
\def\obf{{\bf o}}
\def\pbf{{\bf p}}
\def\qbf{{\bf q}}
\def\rbf{{\bf r}}
\def\sbf{{\bf s}}
\def\tbf{{\bf t}}
\def\ubf{{\bf u}}
\def\vbf{{\bf v}}
\def\wbf{{\bf w}}
\def\xbf{{\bf x}}
\def\ybf{{\bf y}}
\def\zbf{{\bf z}}
\def\rbf{{\bf r}}
\def\xbf{{\bf x}}
\def\ybf{{\bf y}}
\def\Abf{{\bf A}}
\def\Bbf{{\bf B}}
\def\Cbf{{\bf C}}
\def\Dbf{{\bf D}}
\def\Ebf{{\bf E}}
\def\Fbf{{\bf F}}
\def\Gbf{{\bf G}}
\def\Hbf{{\bf H}}
\def\Ibf{{\bf I}}
\def\Jbf{{\bf J}}
\def\Kbf{{\bf K}}
\def\Lbf{{\bf L}}
\def\Mbf{{\bf M}}
\def\Nbf{{\bf N}}
\def\Obf{{\bf O}}
\def\Pbf{{\bf P}}
\def\Qbf{{\bf Q}}
\def\Rbf{{\bf R}}
\def\Sbf{{\bf S}}
\def\Tbf{{\bf T}}
\def\Ubf{{\bf U}}
\def\Vbf{{\bf V}}
\def\Wbf{{\bf W}}
\def\Xbf{{\bf X}}
\def\Ybf{{\bf Y}}
\def\Zbf{{\bf Z}}
\def\Ac{{\cal A}}
\def\Bc{{\cal B}}
\def\Cc{{\cal C}}
\def\Dc{{\cal D}}
\def\Ec{{\cal E}}
\def\Fc{{\cal F}}
\def\Gc{{\cal G}}
\def\Hc{{\cal H}}
\def\Ic{{\cal I}}
\def\Jc{{\cal J}}
\def\Kc{{\cal K}}
\def\Lc{{\cal L}}
\def\Mc{{\cal M}}
\def\Nc{{\cal N}}
\def\Oc{{\cal O}}
\def\Pc{{\cal P}}
\def\Qc{{\cal Q}}
\def\Rc{{\cal R}}
\def\Sc{{\cal S}}
\def\Tc{{\cal T}}
\def\Uc{{\cal U}}
\def\Vc{{\cal V}}
\def\Wc{{\cal W}}
\def\Xc{{\cal X}}
\def\Yc{{\cal Y}}
\def\Zc{{\cal Z}}

\def\0bf{\mathbf{0}}
\def\1bf{\mathbf{1}}

\def\lambdabf{\boldsymbol{\lambda}}
\def\mubf{\boldsymbol{\mu}}
\def\gammabf{\boldsymbol{\gamma}}
\def\xibf{\boldsymbol{\xi}}
\def\sigmabf{\boldsymbol{\sigma}}
\def\zetabf{\boldsymbol{\zeta}}

\def\Lambdabf{\boldsymbol{\Lambda}}
\def\Mubf{\boldsymbol{\Mu}}
\def\Gammabf{\boldsymbol{\Gamma}}
\def\Xibf{\boldsymbol{\Xi}}
\def\Sigmabf{\boldsymbol{\Sigma}}
\def\Zetabf{\boldsymbol{\Zeta}}

\newcommand\independent{\protect\mathpalette{\protect\independenT}{\perp}}
\def\independenT#1#2{\mathrel{\rlap{$#1#2$}\mkern2mu{#1#2}}}

\newcommand{\bsf}[1]{\textsf{\textbf{#1}}}
\newcommand{\lbsf}[1]{\textsf{\large  \textbf{#1}}}
\newcommand{\Lbsf}[1]{\textsf{\Large  \textbf{#1}}}
\newcommand{\hbsf}[1]{\textsf{\huge  \textbf{#1}}}

\newcommand{\myminipage}[3]{\begin{minipage}[#1]{#2}{#3} \end{minipage}}
\newcommand{\sbs}[4]{\myminipage{c}{#1}{#3} \hfill
\myminipage{c}{#2}{#4}}

\newcommand{\myfig}[2]{\centerline{\psfig{figure=#1,width=#2,silent=}}}
\newcommand{\myfigh}[2]{\centerline{\psfig{figure=#1,height=#2,silent=}}}
\newcommand{\myfigwh}[3]{\centerline{\psfig{figure=#1,width=#2,height=#3,silent=}}}
\newcommand{\beqa}{\begin{eqnarray}}
\newcommand{\eeqa}{\end{eqnarray}}
\newcommand{\beqan}{\begin{eqnarray*}}
\newcommand{\eeqan}{\end{eqnarray*}}
\newcommand{\dst}[1]{\displaystyle{ #1 }}
\newcounter{pf}
\newcommand{\smax}[1] { \bar \sigma \left( #1 \right) }
\newcommand{\Rn}{{\mathbb R}^n}
\newcommand{\R}{{\mathbb R}}
\newcommand{\C}{{\mathbb C}}
\newcommand{\Rm}{\mathbb{R}^m}
\newcommand{\Rmn}{\mathbb{R}^{m \times n}}
\newcommand{\Rpq}{\mathbb{R}^{p \times q}}
\newcommand{\Cn}{\mathbb{C}^n}
\newcommand{\Cm}{\mathbb{C}^m}
\newcommand{\Cnn}{\mathbb{C}^{n \times n}}
\newcommand{\Cmn}{\mathbb{C}^{m \times n}}
\newcommand{\ip}[1]{\left\langle #1 \right\rangle}
\newcommand{\rank}{\mbox{rank}}
\newcommand{\Span}{\mbox{\rm Span }}
\newcommand{\Trace}{\mbox{\rm Tr }}
\newcommand{\Spec}{\mbox{\rm Spec }}
\newcommand{\vectornorm}[1]{\left\|#1\right\|}

\newcommand{\pd}[2]{\frac{\partial #1}{\partial #2}}
\newcommand{\ppd}[3]{\frac{\partial^2 #1}{\partial #2 \partial #3}}

\newcommand{\thtilde}{\tilde{\theta}}
\newcommand{\thnom}{\theta^\circ}
\newcommand{\thopt}{\theta^{\mbox{\small opt}}}
\newcommand{\thhat}{{\hat{\theta}}}
\newcommand{\Tho}{\Theta^\circ}
\newcommand{\tho}{\theta^\circ}
\newcommand{\np}{{n_p}}

\newcommand{\ii}{{[i]}}
\newcommand{\II}{{[i+1]}}
\newcommand{\iii}{{[ii]}}
\newcommand{\jj}{{[j]}}
\newcommand{\kk}{{[k]}}
\newcommand{\thi}{{\theta^\ii}}
\newcommand{\thI}{{\theta^\II}}
\newcommand{\di}{{d^\ii}}
\newcommand{\gi}{{g^\ii}}
\newcommand{\Hi}{{\HH^\ii}}
\newcommand{\thK}{\theta^{(k+1)}}
\newcommand{\gk}{{g^{(k)}}}
\newcommand{\Hk}{{{\cal H}^{(k)}}}

\newcommand{\bfdelta}{{\bf \Delta}}

\newcommand{\pr}[1]{\mathbb{P} \left\{ #1 \right\}}
\newcommand{\prC}[1]{\mathbb{P}_0 \left\{ #1 \right\}}
\newcommand{\prN}[1]{\mathbb{P}_0^N \left\{ #1 \right\}}
\newcommand{\Exp}[1]{\exp \left\{ #1 \right\}} % exponential
\newcommand{\gaussian}[1]{\mathbb{N} \left( #1 \right)}
\newcommand{\uniform}[1]{\mathbb{U} \left[ #1 \right]}
\newcommand{\exponential}[1]{\mathbb{E} \left[ #1 \right]}
\newcommand{\EXP}[1]{\EEXP \left[ #1 \right]} % expectation
\newcommand{\EEXP}{\mbox{\bsf{E}}} % expectation
\newcommand{\Prob}[1]{\mbox{{\sf Pr}} \left(#1 \right)}
\newcommand{\convas}{\stackrel{as}{\longrightarrow}}
\newcommand{\convinp}{\stackrel{p}{\longrightarrow}}
\newcommand{\convind}{\stackrel{d}{\longrightarrow}}
\newcommand{\convqm}{\stackrel{qm}{\longrightarrow}}
\newcommand{\sss}[1]{{_{#1}}}
\newcommand{\density}[2]{p_{_{_{#1}}}\!\!\left(#2 \right)} % density
\newcommand{\distro}[2]{P_{_{_{#1}}}\!\!\left(#2 \right)} % distribution
\newcommand{\rxx}[1]{R_{_{#1}}\!} % auto and cross correlations
\newcommand{\sxx}[1]{S_{_{#1}}} % power spectral densities
\newcommand{\cov}[1]{\Lambda_{_{#1}}} % covariance matrices
\newcommand{\mean}[1]{m_{_{#1}}} % mean
\newcommand{\LS}[1]{\hat{#1}_{_{LS}}} % least-squares estimate
\newcommand{\MV}[1]{\hat{#1}_{_{MV}}} % minimum variance estimate
\newcommand{\LMV}[1]{\hat{#1}_{_{LMV}}} % linear minimum variance estimate
\newcommand{\ML}[1]{\hat{#1}_{_{ML}}} % maximum likelihood estimate

\renewcommand{\arraystretch}{0.9}
\newcommand{\bmat}[1]{ \begin{bmatrix} #1 \end{bmatrix}}
\newcommand{\mat}[1]{ \left[ \begin{array}{cccccccc} #1 \end{array}
\right] }
\newcommand{\smallmat}[1]{\small{\mat{#1}}}
\newcommand{\sysblk}[4]{\begin{array}{c|cccc}#1&#2\\ \hline#3&#4
\end{array}}
\newcommand{\sysmat}[4]{\left[\sysblk{#1}{#2}{#3}{#4}\right]}
\newcommand{\SGeq}{\succ}
\newcommand{\SLeq}{\prec}
\newcommand{\Geq}{\succeq}
\newcommand{\Leq}{\preceq}

\newcommand{\Bset}{\mathbb{B}}
\newcommand{\Cset}{\mathbb{C}}
\newcommand{\Fset}{\mathbb{F}}
\newcommand{\Mset}{\mathbb{M}}
\newcommand{\Nset}{\mathbb{N}}
\newcommand{\Pset}{\mathbb{P}}
\newcommand{\Qset}{\mathbb{Q}}
\newcommand{\Rset}{\mathbb{R}}
\newcommand{\Sset}{\mathbb{S}}
\newcommand{\Tset}{\mathbb{T}}
\newcommand{\Uset}{\mathbb{U}}
\newcommand{\Vset}{\mathbb{V}}
\newcommand{\Wset}{\mathbb{W}}
\newcommand{\Zset}{\mathbb{Z}}

\newcommand{\Ical}{{\cal I}}
\newcommand{\Acal}{{\cal A}}
\newcommand{\Bcal}{{\cal B}}
\newcommand{\Ccal}{{\cal C}}
\newcommand{\Dcal}{{\cal D}}
\newcommand{\Ecal}{{\cal E}}
\newcommand{\Fcal}{{\cal F}}
\newcommand{\Gcal}{{\cal G}}
\newcommand{\Hcal}{{\cal H}}
\newcommand{\Kcal}{{\cal K}}
\newcommand{\Lcal}{{\cal L}}
\newcommand{\Mcal}{{\cal M}}
\newcommand{\Ncal}{{\cal N}}
\newcommand{\Pcal}{{\cal P}}
\newcommand{\Qcal}{{\cal Q}}
\newcommand{\Rcal}{{\cal R}}
\newcommand{\Scal}{{\cal S}}
\newcommand{\Tcal}{{\cal T}}
\newcommand{\Wcal}{{\cal W}}
\newcommand{\Ucal}{{\cal U}}
\newcommand{\Vcal}{{\cal V}}
\newcommand{\Xcal}{{\cal X}}
\newcommand{\Zcal}{{\cal Z}}

\newcommand{\FF}{{\bf F}}
\newcommand{\GG}{{\bf G}}
\newcommand{\HH}{{\bf H}}
\newcommand{\LL}{{\bf L}}
\newcommand{\NN}{{\bf N}}
\newcommand{\MM}{{\bf M}}
\newcommand{\PP}{{\bf P}}
\newcommand{\QQ}{{\bf Q}}
\newcommand{\RR}{{\bf R}}
\renewcommand{\SS}{{\bf S}}
\newcommand{\TT}{{\bf T}}
\newcommand{\VV}{{\bf V}}
\newcommand{\WW}{{\bf W}}

\newcommand{\thk}{\theta^{(k)}}
\newcommand{\thb}{\theta^{\rm opt}}
\newcommand{\alb}{\alpha^{\rm opt}}
\newcommand{\dk}{d^{(k)}}
\newcommand{\Hinf}{{\cal H}_\infty}
\newcommand{\Htwo}{{\cal H}_2}

\renewcommand{\arraystretch}{1.1}

\newcommand{\red}[1]{{\color{red} #1}}
\newcommand{\blue}[1]{{\color{Blue} #1}}

\newcounter{l1}
\newcounter{l2}
\newcounter{l3}
\newcommand{\bdotlist}{\begin{list}{$\bullet$}{}}
\newcommand{\bboxlist}{\begin{list}{$\Box$}{}}
\newcommand{\bbboxlist}{\begin{list}{\raisebox{.005in}{{\tiny
$\blacksquare$ \ \ }}}{}}
\newcommand{\bdashlist}{\begin{list}{$-$}{} }
\newcommand{\blist}{\begin{list}{}{} }
\newcommand{\barablist}{\begin{list}{\arabic{l1}}{\usecounter{l1}}}
\newcommand{\balphlist}{\begin{list}{(\alph{l2})}{\usecounter{l2}}}
\newcommand{\bAlphlist}{\begin{list}{\Alph{l2}.}{\usecounter{l2}}}
\newcommand{\bdiamlist}{\begin{list}{$\diamond$}{}}
\newcommand{\bromalist}{\begin{list}{(\roman{l3})}{\usecounter{l3}}}

\newcommand{\thm}[1]{\noindent \begin{theorem} #1   \end{theorem}}
\newcommand{\prop}[1]{\begin{proposition} #1 \end{proposition}}
\newcommand{\lem}[1]{\begin{lemma} #1  \hfill $\blacksquare$ \end{lemma}}
\newcommand{\ex}[1]{\begin{example} {\rm #1} \end{example}}
\newcommand{\prf}[1]{ \noindent {\em Proof:} \, #1 \hfill $\blacksquare$}
\newcommand{\rem}[1]{\begin{remark} {\rm #1} \hfill $\Box$ \end{remark}}
\newcommand{\defn}[1]{\begin{definition} {\rm #1 } \end{definition}}
\newcommand{\prob}[1]{\begin{exercise} {\rm  #1 } \end{exercise}}
\newcommand{\cor}[1]{\begin{corollary}   #1  \end{corollary}}

\usepackage{pgfplots}
\pgfplotsset{compat=newest}
\pgfplotsset{every tick label/.append style={font=\small}}

\IEEEoverridecommandlockouts
\begin{document}

\title{\huge Risk-Sensitive Learning and Pricing for Demand Response}

\author{Kia Khezeli \qquad Eilyan Bitar  
\thanks{Supported in part by NSF grant ECCS-1351621, NSF grant CNS-1239178, NSF grant IIP-1632124, US DoE under the CERTS initiative, and the Simons Institute for the Theory of Computing.}
\thanks{Kia Khezeli and  Eilyan Bitar  are with the School of Electrical and Computer Engineering, Cornell University, Ithaca, NY, 14853, USA.  Emails: {\tt\small \{kk839, eyb5\}@cornell.edu}}}

\maketitle

%%%%%%%%%%%%%%%%%%%%%%%%%%%%%%%%%%%%%%%%%%%%%%%%%%%%%%%%%%%%%%%%%%%%%%%%%%%%%%
%%%%%%%%%%%%%%%%%%%%%%%%%%%%%%%%%%%%%%%%%%%%%%%%%%%%%%%%%%%%%%%%%%%%%%%%%%%%%%

\begin{abstract}

We consider the setting in which an  electric power utility seeks to curtail its peak electricity demand by offering a fixed group of customers a uniform price for reductions in consumption relative to their predetermined baselines. The underlying demand curve, which describes the aggregate reduction in consumption in response to the offered price, is assumed to be affine and subject to unobservable random shocks. Assuming that both the parameters of the demand curve and the distribution of the random shocks are initially unknown to the utility, we investigate the extent to which the utility might dynamically adjust its offered prices to maximize its cumulative \emph{risk-sensitive payoff} over a finite number of $T$ days. In order to do so effectively, the utility must design its pricing policy to balance the tradeoff between the need to learn the unknown demand model (exploration) and maximize its payoff (exploitation) over time. In this paper, we propose such a pricing policy, which is shown to exhibit an expected payoff loss over $T$ days that is at most $O(\sqrt{T}\log(T))$, relative to an oracle pricing policy that knows the underlying demand model. Moreover, the proposed pricing policy is shown to yield a sequence of prices that converge to the oracle optimal prices  in the mean square sense.

\end{abstract} 
\section{Introduction}

The ability to implement  residential \emph{demand response} (DR) programs at scale has the potential to substantially improve the efficiency and reliability of electric power systems. In the following paper, we consider a class of DR programs in  which an electric power utility seeks to elicit a reduction in the aggregate electricity demand of a fixed group of customers, during peak demand periods. The class of DR programs we consider rely on non-discriminatory, price-based incentives for demand reduction. That is to say, each participating customer is remunerated for her reduction in electricity demand according to a uniform price determined by the utility. 

There are several challenges a utility faces in implementing such programs, the most basic of which is the  prediction of  how customers will adjust their aggregate demand in response to different prices -- the so-called aggregate demand curve.  The extent to which customers are willing to forego consumption,  in exchange for monetary compensation, is  contingent on variety of idiosyncratic and stochastic factors -- the majority  of which are initially unknown or not directly measurable  by the utility. The utility must, therefore, endeavor to learn the behavior of customers over time through observation of aggregate demand reductions in response to  its offered prices for DR. At the same time, the utility must set its prices for DR in such a manner as to promote increased earnings over time. As we will later establish, such tasks are inextricably linked, and give rise to a trade-off between \emph{learning} (exploration) and \emph{earning} (exploitation) in pricing demand response over time.

\emph{Contribution and Related Work:} \
We consider the setting in which the electric power utility is faced with a demand curve that is affine in price, and subject to unobservable, additive random shocks. Assuming that both the parameters of the demand curve and the distribution of the random shocks are initially unknown to the utility, we investigate the extent to which the utility might dynamically adjust its offered prices for demand curtailment to maximize its cumulative risk-sensitive payoff over a finite number of $T$ days. We define the utility's payoff on any given day as the largest return the utility is guaranteed to receive with probability no less than $1- \alpha$. Here, $\alpha \in (0,1)$ encodes the utility's sensitivity to risk. In this paper, we propose a causal pricing policy, which resolves the trade-off between the utility's need to learn the underlying demand model and maximize its cumulative risk-sensitive payoff over time. More specifically,  the proposed pricing policy  is shown to exhibit an expected payoff loss over $T$ days -- relative to an oracle that knows the underlying demand model -- which is at most $O(\sqrt{T}\log(T))$. Moreover, the proposed pricing policy is shown to yield a sequence of offered prices, which converges to the sequence of oracle optimal prices  in the mean square sense. 

There is a related stream of literature in operations research and adaptive control \cite{besbes2015surprising,den2013simultaneously,keskin2014dynamic,lai1982iterated,den2015dynamic}, which considers a similar setting in which a monopolist  endeavors to sell a product over multiple time periods -- with the aim of maximizing its cumulative expected revenue --  when the underlying demand curve (for that product) is unknown and subject to exogenous shocks. What distinguishes our formulation from this prevailing literature is the  explicit treatment of risk-sensitivity in the optimization criterion we consider, and the subsequent need to design pricing policies that  not only learn the underlying demand curve, but also learn the shock distribution. 

Focusing explicitly on demand response applications, there  are  several related papers in the literature, which formulate the problem of eliciting demand response under uncertainty within the framework of multi-armed bandits \cite{taylor2014index,kalathil2015online,jain2014multiarmed, wang2014adaptive}. In this setting, each arm represents a customer or a class of customers. Taylor and Mathieu \cite{taylor2014index} show that, in the absence of exogenous shocks on load curtailment, the optimal policy is indexable. Kalathil and Rajagopal \cite{kalathil2015online} consider a similar multi-armed bandit setting in which a customer's load curtailment is subject to an exogenous shock, and attenuation due to fatigue resulting from repeated requests for reduction in demand over time. They propose a policy, which guarantees that the $T$-period regret is bounded from above by $O(\sqrt{T\log T})$. There is a related stream of literature, which treats the problem of pricing demand response under uncertainty using techniques from online learning \cite{gomez2012learning, jia2014online, neill2010residential, soltani2015real}. Perhaps closest to the setting considered in this paper, Jia et al. \cite{jia2014online} consider the problem of pricing demand response when the underlying demand function is unknown, affine, and subject to normally distributed random shocks. With the aim of maximizing the utility's expected surplus, they propose a stochastic approximation-based pricing policy, and establish an upper bound on the $T$-period regret that is of the order $O(\log T)$. There is another stream of literature, which considers an auction-based approach to the procurement of demand response \cite{bitar2013incentive,bitar2016deadline,lin2014forward, mohsenian2010autonomous, saad2012game,xu2015demand,tavafoghi2014optimal}. In such settings, the primary instrument for analysis is game-theoretic in nature.

\emph{Organization:} \ The rest of the paper is organized as follows. In Section \ref{sec:model}, we develop the demand model and  formulate the utility's  pricing problem for demand response. In Section \ref{sec:learning}, we outline a scheme for demand model learning. In  Section \ref{sec:pricing}, we propose a pricing policy and analyze its performance. We investigate the behavior of the proposed pricing policy with a numerical case study in Section \ref{sec:numerical}. All mathematical proofs are presented in the Appendix to the paper.  
\section{Model} \label{sec:model}

\subsection{Responsive Demand Model}

We consider a class of demand response (DR) programs in  which an electric power utility seeks to elicit a reduction in peak electricity demand from a fixed group of $N$ customers over multiple time periods (e.g., days) indexed by $t = 1,2, \dots$. The class of DR programs we consider rely on uniform price-based incentives for demand reduction.\footnote{This class of DR programs falls within the more general category of programs that rely on \emph{peak time rebates} (PTR) as incentives for demand reduction \cite{faruqui2009power}.} Specifically, prior to each time period $t$, the utility broadcasts a single price $p_t \geq 0$ (\$/kWh), to which each participating customer $i$ responds with a reduction in demand $D_{it}$ (kWh) -- thus entitling  customer $i$ to receive a payment in the amount of  $p_t D_{it}$.\footnote{A customer's reduction in demand is measured against a predetermined baseline. The question as to how such baselines might be reliably inferred is a challenging and active area of research \cite{chao2011demand, chelmis2015curtailment, conedison2013energy, coughlin2009statistical, muthirayan2016mechanism}. Expanding our model to make endogenous the calculation of customer baselines is left as a direction for future research.}

We model the response of each customer $i$ to the posted price $p_t$ at time $t$ according to a linear demand function given by 
\begin{align*}
D_{it} = a_ip_t+b_i+\varepsilon_{it}, \quad \text{for} \  \  i = 1, \dots, N,
\end{align*} 
where $a_i \in \Rset$ and $b_i \in \Rset$ are model parameters \emph{unknown to the utility}, and  $\varepsilon_{it}$ is an unobservable demand shock, which we model as a random variable with zero mean.\footnote{We note that the assumption that $\varepsilon_{it}$ be zero-mean is without loss of generality.} \emph{Its distribution is also unknown to the utility}. We define the aggregate response of customers at time $t$ as $D_t := \sum_{i=1}^N D_{it}$, which satisfies 
\begin{align} \label{eq:aggdemand}
D_t = a p_t + b + \varepsilon_t.
\end{align}
Here, the aggregate model parameters  and shock are defined as $a := \sum_{i=1}^N a_i$,  $b := \sum_{i=1}^N b_i$, and  $\varepsilon_t := \sum_{i=1}^N \varepsilon_{it}$. To simplify notation in the sequel,  we  write  the deterministic component of aggregate demand as $\lambda(p, \theta) := ap + b$, where $\theta := (a,b)$ denotes the aggregate demand function parameters. 

We assume throughout the paper that $a \in\left[\underline{a},\overline{a}\right]$   and  $b \in\left[0,\overline{b}\right]$,  where  the model parameter bounds are assumed to be known and satisfy $0  < \underline{a} \leq \overline{a} < \infty$ and  $ 0  \leq \overline{b} < \infty$.  Such assumptions are natural, as they ensure that the price elasticity of aggregate demand is strictly positive and bounded, and that reductions in aggregate demand are guaranteed to be nonnegative in the absence of demand shocks. We also assume that the sequence of shocks $\{\varepsilon_t\}$ are independent and identically distributed random variables, in addition to the following technical assumption.

\begin{assumptio} \label{ass:bilip}  The aggregate demand shock  $\varepsilon_t$ has a bounded range $[\underline{\varepsilon}, \overline{\varepsilon}]$, and a cumulative distribution function $F$, which is bi-Lipschitz over this range. Namely, there exists a real constant $L \geq 1$, such that for all $x,y\in [\underline{\varepsilon}, \overline{\varepsilon}]$, it holds that
 \begin{align*}
\frac{1}{L}\left| x-y \right| \leq \left| F(x)-F(y) \right| \leq L\left| x-y \right|.
\end{align*}
\end{assumptio}
There is a large family of distributions respecting Assumption \ref{ass:bilip} including uniform and doubly truncated normal distributions.
Moreover, the assumption that the aggregate demand shock takes bounded values is natural, given the inherent physical limitation on the range of values that demand can take.   
And, technically speaking, the requirement  that $F$ be bi-Lipschitz is stated to ensure Lipschitz continuity  of its inverse, which will prove critical to the derivation of our main results. Finally, we note that the electric power utility need not know the parameters specified in Assumption \ref{ass:bilip}, beyond the assumption of their boundedness. 

\begin{remar}[On the Linearity Assumption] While the assumption of linearity in the underlying demand model might appear restrictive at first glance, there are several sensible arguments in support of its adoption. 
First, the assumption of linearity is routinely employed in the revenue management and pricing literature \cite{bertsimas2015data,jia2014online, jia2013day, keskin2014dynamic, talluri2006theory,taylor1974asymptotic}, as it serves to facilitate theoretical  analyses, thereby bringing to light key features of the problem and its solution structure. More practically, if the range of allowable prices is sufficiently limited, then it is reasonable to assume that the underlying (possibly nonlinear) demand function is well approximated by an affine function over that range. And, in the specific context of pricing for DR programs, it is reasonable to expect that the electric power utility, being a regulated company, will face restrictions on the range of prices that it can offer to customers. Finally, there are recent results in the revenue management literature \cite{besbes2015surprising}, which demonstrate how the  assumption of a linear demand model might be dynamically \emph{adapted} to  price in environments where the true demand function is nonlinear. The generalization of such techniques to accommodate the risk-sensitive criteria considered in this paper (cf. Equation \eqref{eq:risk}) represents an interesting direction for future research.
\end{remar}

\subsection{Utility Model and Pricing Policies}
We consider a setting in which the utility seeks to reduce its peak electricity demand over multiple days, indexed by $t$. Accordingly, we let $\wholesale_t$ (\$/kWh) denote the  wholesale price of electricity during peak demand hours on day $t$. And, we let $f$ (\$/kWh) denote the retail price of electricity, i.e., the fixed price that customers are charged for their electricity consumption. For the remainder of the paper,  it will be convenient to work with the difference between the wholesale and retail prices of electricity on each day $t$, which we denote by  $c_t:=\wholesale_t-f$. We assume throughout the paper that  $c_t \in [0, \overline{c}]$ for all  days $t$, where $0 \leq \overline{c} < \infty$.\footnote{Implicit in this requirement is the assumption that $f \leq w_t \leq \overline{c} + f$  for all days $t$. The lower bound on $w_t$ implies that the utility will only call for a demand reduction on those days in which the wholesale market manifests in prices that exceed the fixed retail price for electricity. The upper bound on $w_t$ implies the enforcement of a \emph{price cap} in the wholesale market.} In addition,  we assume that $c_t$ is known to the utility  prior to its determination of the  DR price $p_t$ in each period $t$. Upon broadcasting a price $p_t$ to its customer base, and realizing an aggregate demand reduction $D_t$, the utility derives a net reduction in its peak electricity cost in the amount of  $(c_t - p_t) D_t$. Henceforth, we will refer to the net savings $(c_t - p_t) D_t$ as the \emph{revenue} derived by the utility in period $t$.

The utility is assumed to be \emph{sensitive to risk}, in that it would like to set the price for DR in each period $t$ to maximize the revenue it is guaranteed to receive with probability no less than $1-\alpha$. Clearly,  the parameter $\alpha \in (0,1)$ encodes the degree to which the  utility is sensitive to risk. Accordingly, we define the \emph{risk-sensitive revenue} derived by the utility in period $t$ given a posted price $p_t$ as  
\begin{align}  \label{eq:risk}
r_\alpha(p_t):=\sup \left\{x \in \Rset : \mathbb{P}\{(c_t-p_t)D_t \geq x\}\geq 1-\alpha\right\}.
\end{align}
The risk measure specified in \eqref{eq:risk} is closely related to the standard concept of \emph{value at risk}  commonly used in  mathematical finance. Conditioned on a fixed price $p_t$, one can reformulate the expression in \eqref{eq:risk} as
\begin{align} \label{eq:risksimple}
r_\alpha(p_t) = (c_t - p_t)( \lambda(p_t, \theta) + F^{-1}(\alpha)),
\end{align}
where $F^{-1}(\alpha) := \inf \{x \in \Rset :  F(x) \geq \alpha\}$ denotes the $\alpha$-quantile of the random variable $\varepsilon_t$. It is immediate to see from the simplified expression in  \eqref{eq:risksimple} that 
$r_\alpha(p_t)$ is strictly concave in $p_t$.
Let $p_t^*$ denote the \emph{oracle optimal price}, which maximizes the risk-sensitive revenue in period $t$. Namely,
\begin{align*}
p_t^* := \arg \max \{r_\alpha(p_t) : p_t \in [0, c_t]\}.
\end{align*}
The optimal price is readily derived from the corresponding first order  optimality condition, and is given by
\begin{align*}
p_t^* =  \frac{c_t}{2}-\frac{b+F^{-1}(\alpha)}{2a}.
\end{align*}
We define the \emph{oracle risk-sensitive revenue}  accumulated over $T$ time periods as
\begin{align*}
R^*(T) := \sum_{t=1}^T r_{\alpha}(p^*_t).
\end{align*}
The term oracle is used, as $R^*(T)$ equals the maximum risk-sensitive revenue achievable by the utility over $T$ periods if it were to have \emph{perfect knowledge} of the demand model.

In the setting considered in this paper, we assume that both the  demand model parameters $\theta = (a,b)$ and the shock distribution $F$ are \emph{unknown to the utility} at the outset. As a result, the utility must attempt to learn them over time by observing aggregate demand reductions in response to offered prices. Namely, the utility must endeavor to learn the demand model, while simultaneously trying to maximize its risk-sensitive returns over time. As we will later see, such task will naturally  give rise to a trade-off between \emph{learning} (exploration) and \emph{earning} (exploitation) in pricing demand response over time. First, we describe the space of feasible pricing policies.

We assume that, prior to its determination of the DR price in period $t$, the utility has access to the entire history of prices and demand reductions until period $t-1$.  We, therefore, define a \emph{feasible pricing policy} as an infinite  sequence of functions $\pi := (p_1, p_2, \dots)$, where each function in the sequence is allowed to depend only on the past history. More precisely, we require that the function $p_t$ be measurable according to the $\sigma$-algebra generated by the history of past decisions and demand observations $(p_1, \dots, p_{t-1}, D_1, \dots, D_{t-1})$ for all $t \geq 2$, and that $p_1$ be a deterministic constant. The \emph{expected risk-sensitive revenue} generated by a feasible pricing policy $\pi$ over $T$ time periods is defined as
\begin{align*}
R^{\pi}(T) :=   \mathbb{E}^{\pi} \left[ \sum_{t=1}^T r_{\alpha}(p_t) \right],
\end{align*}
where expectation is taken with respect to the demand model \eqref{eq:aggdemand} under the pricing policy $\pi$.

\subsection{Performance Metric} \label{sec:metric}
We evaluate the performance of a feasible pricing policy $\pi$ according to the $T$-period \emph{regret}, which we define as
\begin{align*}
\Delta^{\pi}(T) := R^*(T) - R^{\pi}(T).
\end{align*}
Naturally, pricing policies yielding a small regret are preferred, as the  oracle risk-sensitive revenue $R^*(T)$  stands as an upper bound on the expected risk-sensitive revenue $R^{\pi}(T)$ achievable by any feasible pricing policy $\pi$. Ultimately, we seek a pricing policy whose $T$-period regret is sublinear in the horizon $T$. Such a pricing policy is said to have \emph{no-regret}.
\begin{definitio}[No Regret Pricing] A feasible pricing policy $\pi$ is said to exhibit \emph{no-regret} if $\lim_{T \rightarrow \infty} \Delta^{\pi}(T)/T = 0$.
\end{definitio}
Implicit in the goal of designing a no-regret policy is that the sequence of prices that it generates should converge to the oracle optimal price sequence.
\section{Demand Model Learning} \label{sec:learning} 
Clearly, the ability to price with no-regret will rely centrally on the rate at which the unknown parameters, $\theta$, and  quantile function, $F^{-1}(\alpha)$, can be learned from the market data. In what follows, we describe a basic approach to learning the demand model using the method of least squares estimation.

\subsection{Parameter Estimation}
Given the history of past decisions and demand observations $(p_1, \dots, p_t, D_1, \dots, D_t)$ through period $t$, define the \emph{least squares estimator} (LSE) of $\theta$  as
\begin{align*}
\theta_t := \arg \min \left\{ \sum_{k=1}^t (D_k - \lambda(p_k, \vartheta))^2 \ : \ \vartheta \in \Rset^2 \right\},
\end{align*}
for time periods $t=1, 2, \dots$. The LSE at period $t$ admits an explicit expression of the form
\begin{align} \label{eq:LSE}
\theta_t = \left( \sum_{k=1}^t \bmat{p_k \\ 1} \bmat{p_k \\ 1}^\top \right)^{-1} \left(  \sum_{k=1}^t \bmat{p_k \\ 1} D_k \right),
\end{align}
provided the indicated inverse exists. It will be convenient to define the $2 \times 2$ matrix
\begin{align*}
\mathscr{J}_t := \sum_{k=1}^t \bmat{p_k \\ 1} \bmat{p_k \\ 1}^\top = \bmat{ \sum_{k=1}^t p_k^2 & \sum_{k=1}^t p_k \\ \sum_{k=1}^t p_k & t}.
\end{align*}
Utilizing the definition of the aggregate demand model  \eqref{eq:aggdemand}, in combination with the expression in \eqref{eq:LSE}, one can obtain the following expression for the parameter estimation error:
\begin{align} \label{eq:LSEerror}
\theta_t - \theta = \mathscr{J}_t^{-1} \left(  \sum_{k=1}^t \bmat{p_k \\ 1} \varepsilon_k \right).
\end{align}

\begin{remar}[The Role of Price Dispersion]\label{rem:J}
The expression for the parameter estimation error in \eqref{eq:LSEerror} reveals how consistency of the LSE  is reliant upon the asymptotic spectrum of the matrix $\mathscr{J}_t$. Namely, the minimum eigenvalue of $\mathscr{J}_t$, must grow unbounded with time, in order that the parameter estimation error converge to zero in probability. In \cite[Lemma 2]{keskin2014dynamic}, the authors establish a sufficient condition for such growth. Specifically, they prove that the  minimum eigenvalue of $\mathscr{J}_t$  is bounded from below (up to a multiplicative constant) by the \emph{sum of squared price deviations} defined as
\begin{align*}
J_t: = \sum_{k=1}^t (p_k - \overline{p}_t)^2,
\end{align*}
where $\overline{p}_t: =  (1/t)\sum_{k=1}^t p_k$. The result  is reliant on the assumption that the underlying pricing policy $\pi$ yields a bounded sequence of prices $\{p_t\}$. An important consequence of such a result is that it reveals the explicit role that \emph{price dispersion} (i.e., exploration) plays in facilitating consistent parameter estimation.
\end{remar}

Finally, given the underlying assumption that the unknown model parameters $\theta$ belong to a compact set defined $\Theta := [\underline{a}, \overline{a}] \times [0, \overline{b}]$, one can improve upon the LSE at time $t$ by projecting it onto the set $\Theta$. Accordingly, we define the \emph{truncated least squares estimator} as
\begin{align} \label{eq:TLSE}
\widehat{\theta}_t := \arg \min \left\{  \| \vartheta - \theta_t \|_2    :  \vartheta \in \Theta \right\}.
\end{align}
Clearly, we have that $\| \widehat{\theta}_t - \theta \|_2   \leq \| \theta_t - \theta\|_2$. 
In the following section, we describe an approach to estimating the underlying quantile function using the parameter estimator defined in \eqref{eq:TLSE}.

\subsection{Quantile Estimation}
Building on the parameter estimator specified in Equation \eqref{eq:TLSE}, we construct an estimator of the unknown quantile function $F^{-1}(\alpha)$ according to the empirical quantile function associated with the demand estimation residuals. Namely, in each period $t$, define the sequence of \emph{residuals} associated with the estimator $\widehat{\theta}_t$ as
\begin{align*}
\widehat{\varepsilon}_{k,t} :=  D_k - \lambda(p_k, \widehat{\theta}_t),  
\end{align*}
for $k=1,\dots, t$. Define their \emph{empirical distribution} as 
\begin{align*}
\widehat{F}_{t}(x) : = \frac{1}{t}\sum_{k=1}^{t}\mathds{1}{\left\{\widehat{\varepsilon}_{k,t}\leq x\right\}},
\end{align*}
and their corresponding \emph{empirical quantile function} as $\widehat{F}_{t}^{-1}(\alpha)  :=  \inf \{x \in \Rset : \widehat{F}_{t}(x) \geq \alpha\}$ for all $\alpha \in (0,1)$. 
It will be useful in the sequel to express the empirical quantile function in terms of the order statistics associated with sequence of residuals. Essentially, the \emph{order statistics} $\widehat{\varepsilon}_{(1),t}, \dots,  \widehat{\varepsilon}_{(t),t}$ are defined as a permutation of 
$\widehat{\varepsilon}_{1,t}, \dots,  \widehat{\varepsilon}_{t,t}$ such that $\widehat{\varepsilon}_{(1),t} \leq \widehat{\varepsilon}_{(2),t} \leq  \dots \leq \widehat{\varepsilon}_{(t),t}.$  With this concept in hand, the empirical quantile function can be equivalently expressed as
\begin{align} \label{eq:quantORD}
\widehat{F}_t^{-1}(\alpha) = \widehat{\varepsilon}_{(i),t},
\end{align}
where the index $i$ is chosen such that $\frac{i-1}{t} < \alpha \leq \frac{i}{t}$. It is not hard to see that $i = \lceil t\alpha \rceil$. Using Equation \eqref{eq:quantORD}, one can relate the quantile estimation error to the parameter estimation error according to the following inequality
\begin{align}
&| \widehat{F}_{t}^{-1}(\alpha)  - F^{-1}(\alpha)| \nonumber\\
&\hspace{3em}\leq  | F_{t}^{-1}(\alpha)  - F^{-1}(\alpha) |  + \left(1 +  p_{(i)}^2 \right)^{1/2} \| \widehat{\theta}_t - \theta \|_2  \label{eq:quant},
\end{align}
where ${F}_{t}^{-1}$ is defined as the empirical quantile function  associated with the sequence of demand shocks $\varepsilon_1, \dots, \varepsilon_t$. Their empirical distribution is defined as 
\begin{align*}
{F}_{t}(x) : = \frac{1}{t}\sum_{k=1}^{t}\mathds{1}{\left\{{\varepsilon}_{k}\leq x\right\}}.
\end{align*}

The inequality in \eqref{eq:quant} reveals that consistency of the quantile estimator \eqref{eq:quantORD} is reliant upon consistency of the both the \emph{parameter estimator} and the \emph{empirical quantile function} defined in terms of the sequence of demand shocks. Consistency  of the former is established in Lemma \ref{lem:parameters} under a suitable choice of a pricing policy, which we specify in Equation \eqref{policy:unknown}. Consistency of the latter is clearly independent of the choice of pricing policy. In what follows, we present a bound on the rate of its convergence in probability.

\begin{propositio}
\label{prop:shock quantile}
Let $\mu_1 := 2/(L^2\log(2))$. It holds that
\begin{align}
\mathbb{P}\{| {F}_{t}^{-1}(\alpha)  - F^{-1}(\alpha) | >  \gamma\}&\leq 2\exp(-\mu_1\gamma^2t)\label{eq:dvoretzky}
\end{align}
for all $\gamma>0$ and $t\geq 2$.
\end{propositio}

Proposition \ref{prop:shock quantile} is similar in nature to \cite[Lemma 2]{dvoretzky1956asymptotic}, which provides a bound on the rate at which the empirical distribution function converges to the true cumulative distribution function in probability. The combination of Assumption \ref{ass:bilip} with \cite[Lemma 2]{dvoretzky1956asymptotic}  enables the derivation of the upper bound in Proposition \ref{prop:shock quantile}.

\section{Design of Pricing Policies}  \label{sec:pricing}
Building on the approach to  demand model learning in Section \ref{sec:learning}, we construct a DR pricing policy, which is guaranteed to  exhibit \emph{no-regret}.

\subsection{Myopic Policy} 
We begin with a description of a natural  approach to pricing, which interleaves the model estimation scheme defined in Section  \ref{sec:learning} with a \emph{myopic} approach to pricing. That is to say, at each stage $t+1$, the utility estimates the demand model parameters and quantile function according to \eqref{eq:TLSE} and \eqref{eq:quantORD}, respectively,  and sets the price according to
\begin{align} \label{eq:myopicPrice}
\widehat{p}_{t+1} =  \frac{c_{t+1}}{2}-\frac{\widehat{b}_t+\widehat{F}_t^{-1}(\alpha)}{2\widehat{a}_t}.
\end{align}
Under this pricing policy, the utility essentially treats its model estimate in each period as if it is correct, and disregards the subsequent impact of its choice of price on its ability to accurately estimate the demand model in future time periods. A danger inherent to a myopic approach to pricing such as this is that the resulting price sequence may fail to elicit information from demand at a rate, which is fast enough to enable consistent model estimation. As a result, the model estimates may converge to incorrect values.  Such behavior is well documented in the literature \cite{den2013simultaneously,keskin2014dynamic,lai1982iterated}, and is commonly referred to as \emph{incomplete learning}. In Section \ref{sec:numerical}, we provide a numerical example, which demonstrates the occurrence of incomplete learning under the myopic pricing policy \eqref{eq:myopicPrice}.

\subsection{Perturbed Myopic Policy}
In order to prevent the possibility  of incomplete  learning, we propose a pricing policy that is guaranteed to elicit information from demand at a sufficient rate through carefully designed perturbations to the myopic pricing policy \eqref{eq:myopicPrice}. The pricing policy we propose is defined as
\begin{align}
p_{t+1}=\begin{cases}
\widehat{p}_{t+1},                                     & t \text{ odd}\\
\widehat{p}_{t}+\frac{1}{2}(c_{t+1}-c_{t})+ \rho \delta_{t+1},  & t \text{ even},
\end{cases}
 \label{policy:unknown}
\end{align}
where $\rho \geq 0$ is a user specified positive constant, and $$\delta_t:=  \mbox{sgn}\left(c_{t}-c_{t-1}\right) \cdot t^{-1/4}.$$ We refer to the policy \eqref{policy:unknown} as the \emph{perturbed myopic policy}.\footnote{In defining the sign function, we require that $ \mbox{sgn}(0) = 1$.}  

The perturbed myopic policy differs from the myopic policy  in two important ways. First, the model parameter estimate, $\widehat{\theta}_t$, and quantile estimate, $\widehat{F}_t^{-1}(\alpha)$, are updated at every other time step. Second, to enforce sufficient price exploration, an offset is added to the myopic price at every other time step. Roughly speaking, the sequence of myopic price offsets $\{\rho \delta_t\}$ is chosen to decay at a rate, which is slow enough to ensure consistent model learning, but not so slow  as to preclude a sub-linear growth rate for regret. In Section \ref{sec:mainthm}, we will show that the combination of these features is enough to ensure consistent parameter estimation and a sub-linear growth rate for the $T$-period regret, which is bounded from above by $O(\sqrt{T}\log(T))$.

\begin{remar}[On the Perturbation Order]
We briefly describe the  rationale behind the selection of the  order of the perturbation sequence as $\delta_t  = O(t^{-1/4})$.
First, notice from Equation \eqref{eq:regretNEW} that the regret incurred by any feasible pricing policy is equal to the sum of the  squared pricing errors generated by the policy. Combining this expression with the upper bound   on the absolute pricing error induced by the perturbed myopic policy in \eqref{eq:price dev}, it becomes clear to see the conflicting effects that the perturbation sequence  has on regret. On the one hand,  an increase in the order of the perturbation sequence will tend to reduce the growth rate of regret by increasing the rate at which the parameter estimation error $\| \widehat{\theta}_t - \theta \|_2$ converges to zero. On the other hand, an increase in the order of the perturbation sequence will tend to have the counterproductive effect of increasing the growth rate of regret by increasing the rate at which the deliberate pricing errors $\rho|\delta_t|$ accumulate. A tradeoff, therefore, emerges in selecting the order of the perturbation sequence.
In Appendix \ref{app:regret-unknown}, we show that among all perturbation sequences that are polynomial in $t$, perturbation  sequences of the order $O(t^{-1/4})$ are optimal in the sense of  minimizing the asymptotic order of our upper bound on regret (ignoring logarithmic factors). 
\end{remar}

\section{A Bound on Regret} \label{sec:mainthm}
Given the demand model considered in this paper, one can express the  $T$-period regret as 
\begin{align} \label{eq:regretNEW}
\Delta^{\pi}(T)=a \sum_{t=1}^T \mathbb{E}^{\pi} \left[ (p_t - p_t^*)^2 \right],
\end{align}
under any pricing policy $\pi$. It becomes apparent, upon examination of Equation \eqref{eq:regretNEW}, that the rate at which regret grows is directly proportional to the rate at which pricing errors accumulate. We, therefore, proceed in deriving a bound on the rate at which the  absolute pricing error  $|p_t- p_t^*|$ converges to zero in probability, under the perturbed myopic policy.

First, it is not difficult  to show that, under the perturbed myopic policy \eqref{policy:unknown}, the absolute pricing error incurred in each even time period $t$ is upper bounded by
\begin{align}\label{eq:price deviation}
& |{p}_{t+1}-{p}_{t+1}^*| \\ 
& \hspace{.2in} \leq {\kappa_1} \| \widehat{\theta}_{t-1} - \theta \|_2 \  + \ {\kappa_2} | \widehat{F}_{t-1}^{-1}(\alpha)  - F^{-1}(\alpha) | \  +  \ \rho |\delta_{t+1}|,   \nonumber
\end{align}
where ${\kappa_1} := (\underline{a}^2+(\overline{b}+\overline{\varepsilon})^2)^{1/2}/(2\underline{a}^2) $ and ${\kappa_2} :={1}/({2\underline{a}})$. The pricing error incurred during odd time periods $t$ is similarly bounded, sans the explicit dependency on the myopic price perturbation. The upper bound in \eqref{eq:price deviation} is intuitive as it consists of three terms: the parameter estimation error,  the quantile estimation error, and the myopic price perturbation -- each of which represents a rudimentary source of pricing error.  

One can further refine the upper bound in \eqref{eq:price deviation}, by leveraging on the fact that, under the perturbed myopic policy, the generated sequence of prices is uniformly bounded. That is to say, $|p_t|\leq \overline{p}$ for all time periods $t$, where 
\begin{align*}
\overline{p} :=\frac{1}{2}\max\left\{\overline{c}-\frac{\underline{\varepsilon}}{\underline{a}}\ ,\ \overline{c}-\frac{\underline{\varepsilon}}{\overline{a}}\ ,\ \frac{\overline{b}+\overline{\varepsilon}}{\underline{a}}\right\}.
\end{align*}
Combining this fact with the previously derived upper bound on the quantile estimation error in \eqref{eq:quant}, we have that
\begin{align}
& |{p}_{t+1}-{p}_{t+1}^*|  \label{eq:price dev}  \\
& \hspace{.2in} \leq \kappa_3 \| \widehat{\theta}_{t-1} - \theta \|_2  \ +  \ \kappa_2 | {F}_{t-1}^{-1}(\alpha)  - F^{-1}(\alpha) |  \ + \  \rho |\delta_{t+1}|  , \nonumber
\end{align}
for even time periods $t$, where {$\kappa_3 := \kappa_1 + \kappa_2(1 + \overline{p}^2)^{1/2}$}. 

Consistency of the perturbed myopic policy  depends on the asymptotic behavior of each term in \eqref{eq:price dev}. The price perturbation converges to zero by construction, and consistency of the empirical quantile function is established in Proposition \ref{prop:shock quantile}. The following Lemma establishes a bound on the mean squared parameter estimation error under the perturbed myopic policy \eqref{policy:unknown}.

\begin{lemm}[Consistent Parameter Estimation]
\label{lem:parameters}
There exists a finite positive constant $\mu_2$ such that, under the perturbed myopic policy \eqref{policy:unknown},
\begin{align*}
&\mathbb{E}\left[ \| \widehat{\theta}_t - \theta \|^2 \right]\leq \frac{\mu_2}{\rho^2}\frac{\log(t)}{\sqrt{t}},
\end{align*}
for all $t \geq 3$ and $\rho>0$. 
\end{lemm}

The following Theorem establishes an $O(\sqrt{T}\log(T))$ upper bound on the $T$-period regret.

\begin{theore}[Sub-linear Regret] \label{thm:regret-unknown}
The $T$-period regret incurred by the perturbed myopic policy \eqref{policy:unknown}   satisfies
\begin{align} \label{bound:sigma zero}
\Delta^\pi(T)   \leq C_0 +  C_1 \sqrt{T}\log(T) +  C_2\log(T), 
\end{align}
for all $T\geq 3$. Here, $C_0$, $C_1$, and $C_2$ are finite positive constants.\footnote{We refer the reader to Equations \eqref{eq:C_0} -\eqref{eq:C_2} for the exact specification of the coefficients $C_0$, $C_1$, and $C_2$.}
\end{theore}

In the process of proving Theorem \ref{thm:regret-unknown}, we also show that the perturbed myopic policy generates a sequence of market prices $\{p_t\}$ that converges to the oracle optimal price sequence $\{p_t^*\}$  in the mean square sense. More formally, we have the following corollary.

\begin{corollar}
[Price Consistency]\label{thm:priceconsistency}
The sequence of  prices $\{p_t\}$ generated by the perturbed myopic policy \eqref{policy:unknown}  satisfies
\begin{align*}
\lim_{t\rightarrow\infty}\mathbb{E}\left[ (p_t-p_t^*)^2 \right]=0,
\end{align*}
where $\{p_t^*\}$ denotes the oracle optimal price sequence. 
\end{corollar}

\subsection{The Exploratory Effect of  Wholesale Price Variation}

Thus far in this paper, we have made no assumption on the nature of variation in the sequence of  wholesale electricity  prices $\{\wholesale_t\}$. In particular, all of the previously stated results hold for any sequence of time-varying wholesale electricity prices. This  includes the special case in which  the wholesale price of electricity is constant across time, i.e., $\wholesale_t = \wholesale$ for all time periods $t$. It is, however, natural to inquire as to  how the degree of  variation in the sequence of wholesale prices might impact the performance of the pricing policies considered in this paper. 

First, it is straightforward to see from Equation \eqref{eq:myopicPrice} that variation in the sequence of wholesale prices induces equivalent variation in the sequence of myopic prices. Such variation in the myopic price sequence is most naturally interpreted as a form of \emph{costless exploration}. In the following result, we establish a sufficient condition on the variation of wholesale prices, which eliminates the need for external perturbations to the myopic price sequence (i.e., setting $\rho =0$), while guaranteeing an upper bound on the resulting $T$-period regret that is $O(\log^2 (T))$. 

\begin{theore}[Logarithmic Regret]
\label{thm:regret-log} Assume that there exists a finite positive constant $\sigma > 0$ such that 
\begin{align} \label{ass:var}
\left|w_t-w_{t-1}\right| \geq \sigma,
\end{align}
for all time periods $t$.\footnote{Note that Assumption \eqref{ass:var} in Theorem \ref{thm:regret-log} implies that $\left|c_t-c_{t-1}\right| \geq \sigma$.} It follows that the $T$-period regret incurred by the perturbed myopic policy \eqref{policy:unknown}, with $\rho = 0$,  satisfies
\begin{align}
\Delta^\pi(T)\leq M_0 +\frac{M_2}{\sigma^2} + M_1 \log(T)+\frac{M_2}{\sigma^2}\log^2(T), \label{bound:regret-log}
\end{align}
for all $T \geq 3$. Here, $M_0, M_1$, and $M_2$ are finite positive constants\footnote{We refer the reader to Equations \eqref{eq:M_0} -\eqref{eq:M_2} for the exact specification of the coefficients $M_0, M_1$, and $M_2$.}, which are independent of the parameter $\sigma$.
\end{theore}

Several comments are in order. First, under the additional assumption of persistent wholesale price variation \eqref{ass:var}, we establish in Theorem \ref{thm:regret-log} an improvement upon the original order of regret stated in Theorem \ref{thm:regret-unknown} from $O(\sqrt{T}\log(T))$ to $O(\log^2 (T))$. However, as one might expect, the magnitude of the upper bound on regret in \eqref{bound:regret-log} scales in a manner that is inversely proportional to $\sigma^2$. As a result,  the upper bound on the $T$-period regret goes to infinity as $\sigma$ goes to zero, and, therefore, provides little useful information when $\sigma$ is small.
\section{Case Study}\label{sec:numerical}

We conduct a numerical analysis to compare the performance of the myopic policy \eqref{eq:myopicPrice} against the perturbed myopic policy \eqref{policy:unknown} over a time horizon of $T=10^4$. We set the tuning parameter $\rho=0.19$.  We  consider the setting in which  there are $N=1000$ customers participating in the DR program. For each customer $i$, we select $a_i$ uniformly at random from the interval $[0.04,0.20]$, and independently select $b_i$ according an exponential distribution  (with mean equal to $0.01$) truncated over  interval $[0,0.1]$.  Parameters are drawn independently across customers.\footnote{It is worth noting that the range of parameter values $a_i \in [0.04,0.20]$ considered in this numerical study  is consistent with the range of  demand price elasticities observed in several real-time pricing programs conducted in the United States \cite{qdr2006benefits,faruqui2010household}.} For  each customer $i$, we take the demand shock to be distributed according to a normal distribution with zero-mean and standard deviation  equal to $0.04$,  truncated over the interval $[-0.4,0.4]$. We consider a utility with risk sensitivity equal to $\alpha=0.1$. In other words, the utility seeks to maximize the revenue it is guaranteed to receive with probability no less than 0.9. Finally, we set the retail price of electricity to $f=0.17$ (\$/kWh), and set the wholesale price of electricity  to $\wholesale_t = 1.67$ (\$/kWh) for all days  $t$. Such values are consistent with the average residential retail  and peak wholesale prices of electricity in the state of New York in 2016 \cite{eia1electric,NYISO}.

\subsection{Discussion}

\begin{figure*}[t]
\centering
\captionsetup{justification=raggedright,singlelinecheck=false}
\subfloat[Sequences of demand parameter $\widehat{a}_t$.]{\label{fig:a}\fbox{\includegraphics[width=0.67\columnwidth]{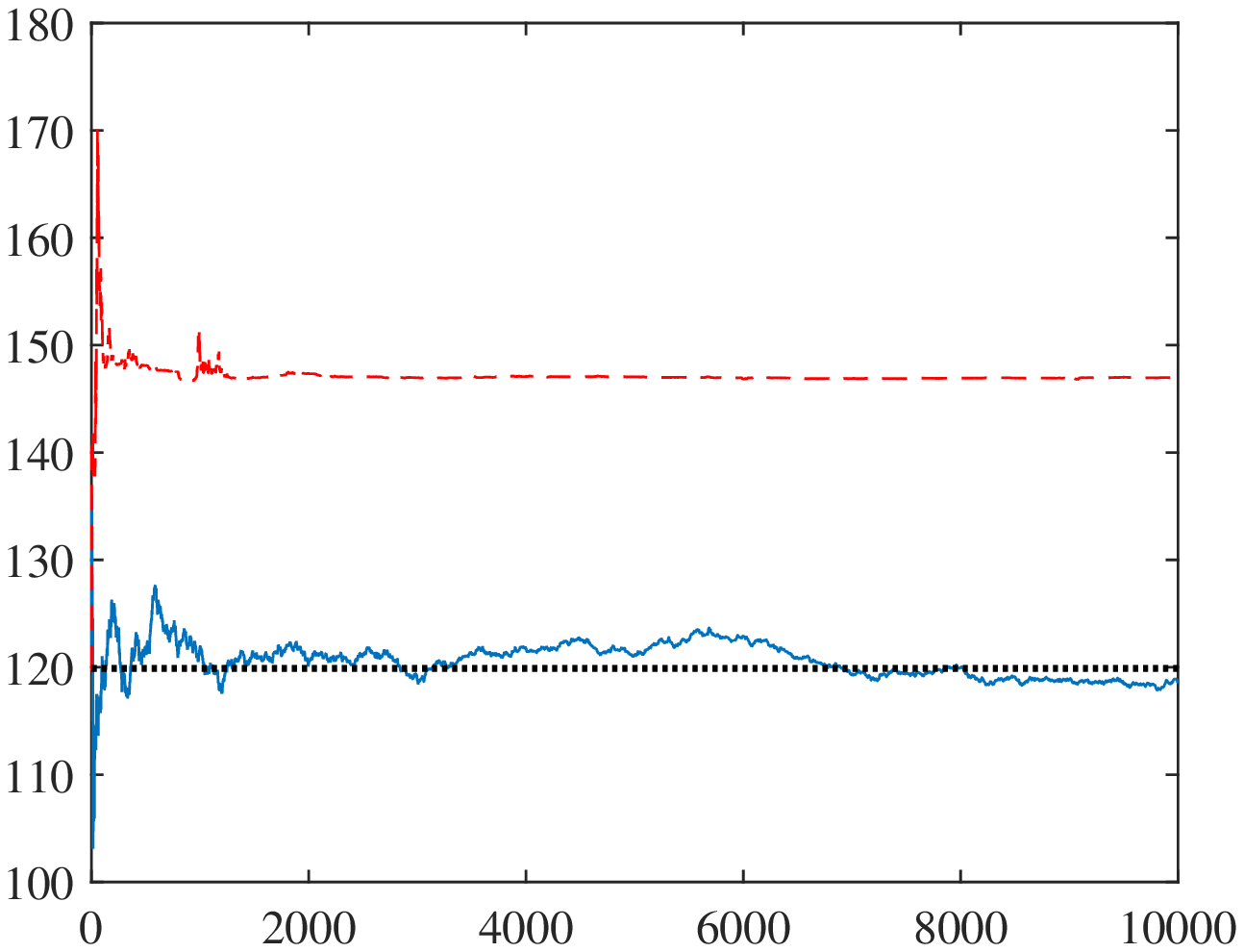}}} 
\subfloat[Sequences of demand parameter $\widehat{b}_t$.]{\label{fig:b}\fbox{\includegraphics[width=0.67\columnwidth]{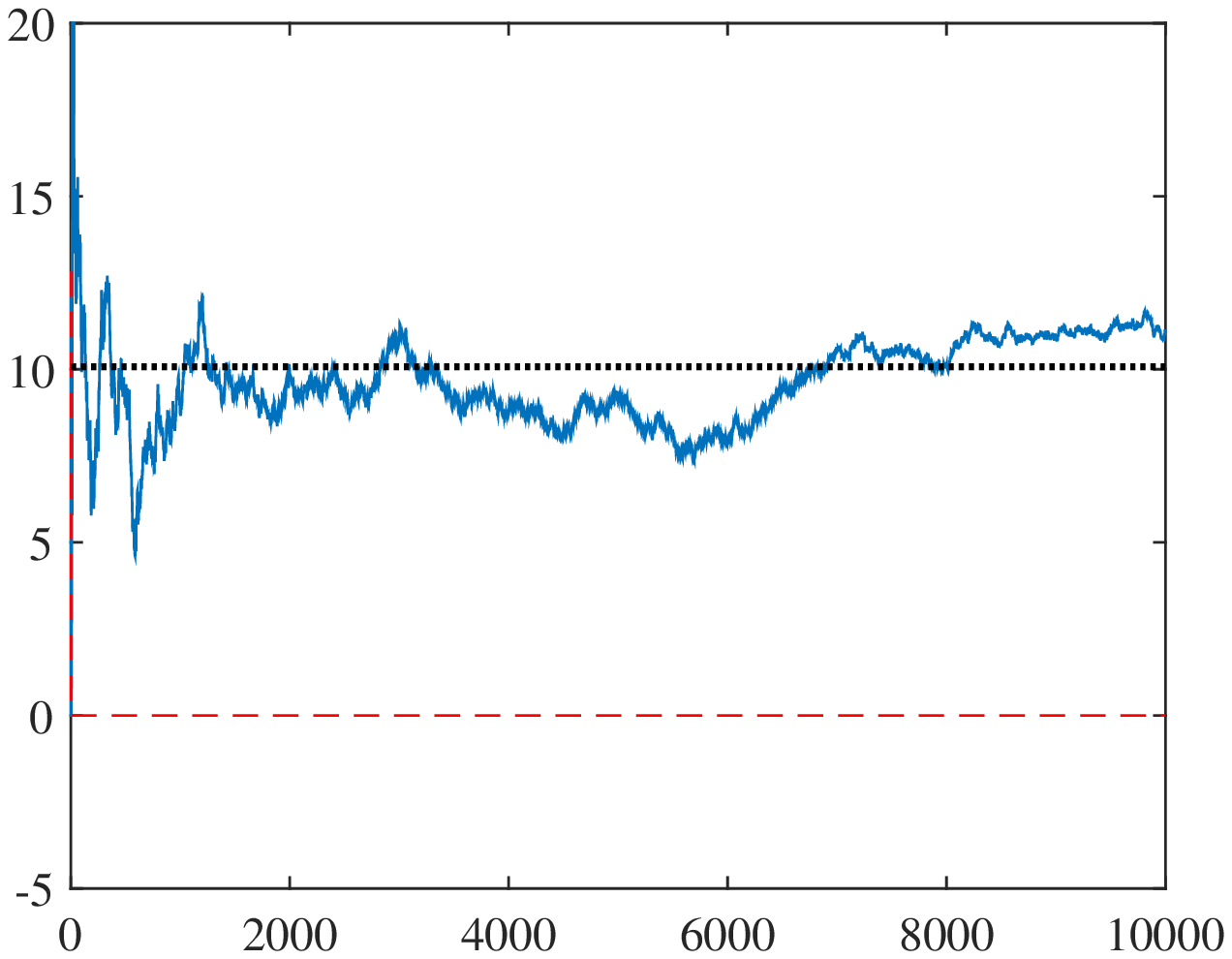}}} 
\subfloat[Sequences of quantile function $\widehat{F}_t^{-1}(\alpha)$.]{\label{fig:Q}\fbox{\includegraphics[width=0.67\columnwidth]{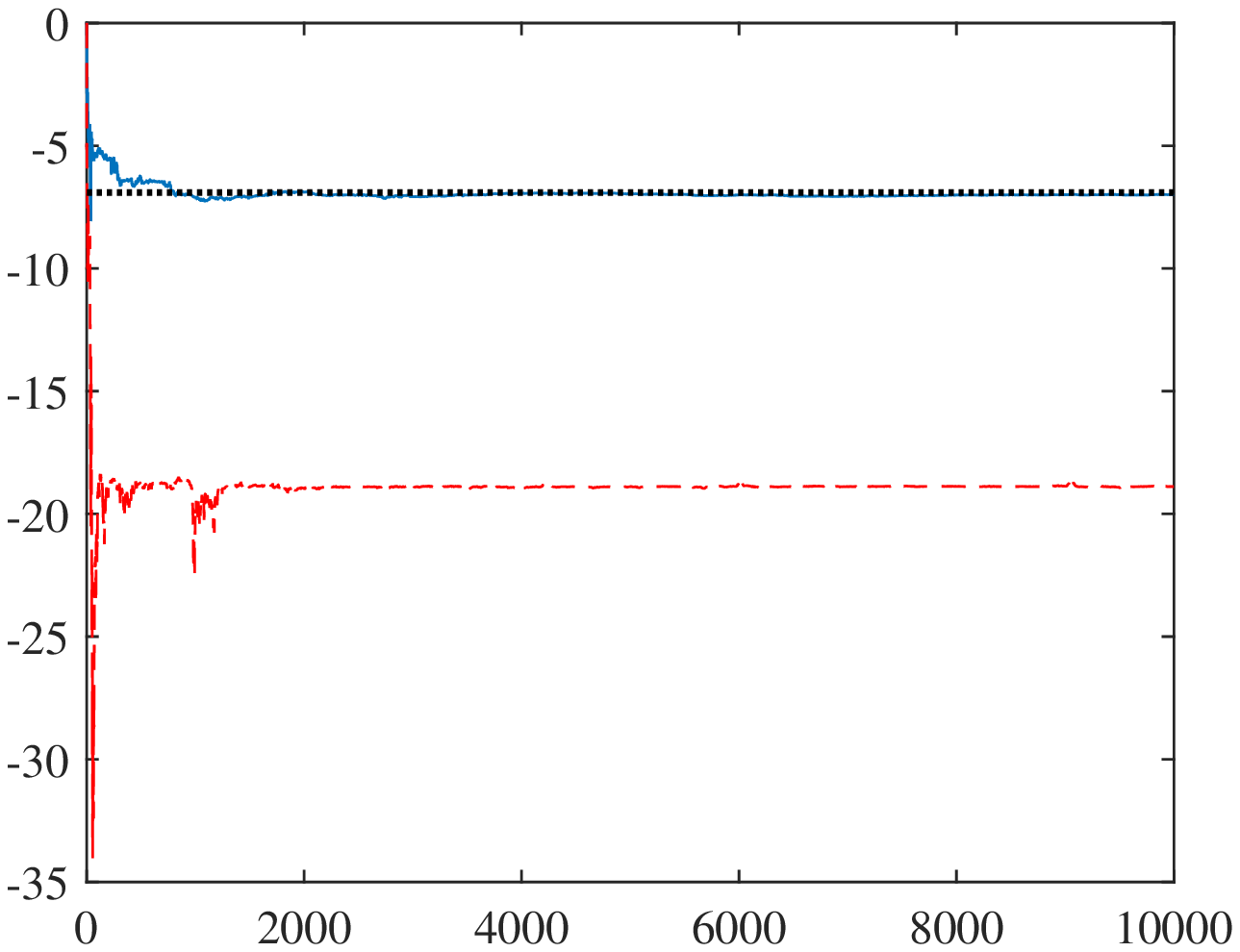}}}
\caption{ (a)-(b) Sample paths of the parameter estimates, and (c) sample path of the shock quantile estimates under the \emph{myopic policy} (\red{\hdashrule[0.5ex]{7mm}{1pt}{2.5pt}}), the \emph{perturbed myopic policy} ({\color[rgb]{0,0.4470,0.7410}{\hdashrule[0.5ex]{7mm}{1pt}{}}}), and the \emph{oracle policy} (\hdashrule[0.5ex]{7mm}{1pt}{1pt}).}\label{fig:par}
\end{figure*}

\begin{figure*}[t]
\centering
\captionsetup{justification=raggedright,singlelinecheck=false}
\subfloat[Sequences of posted prices.]{\label{fig:p}\fbox{\includegraphics[width=0.67\columnwidth]{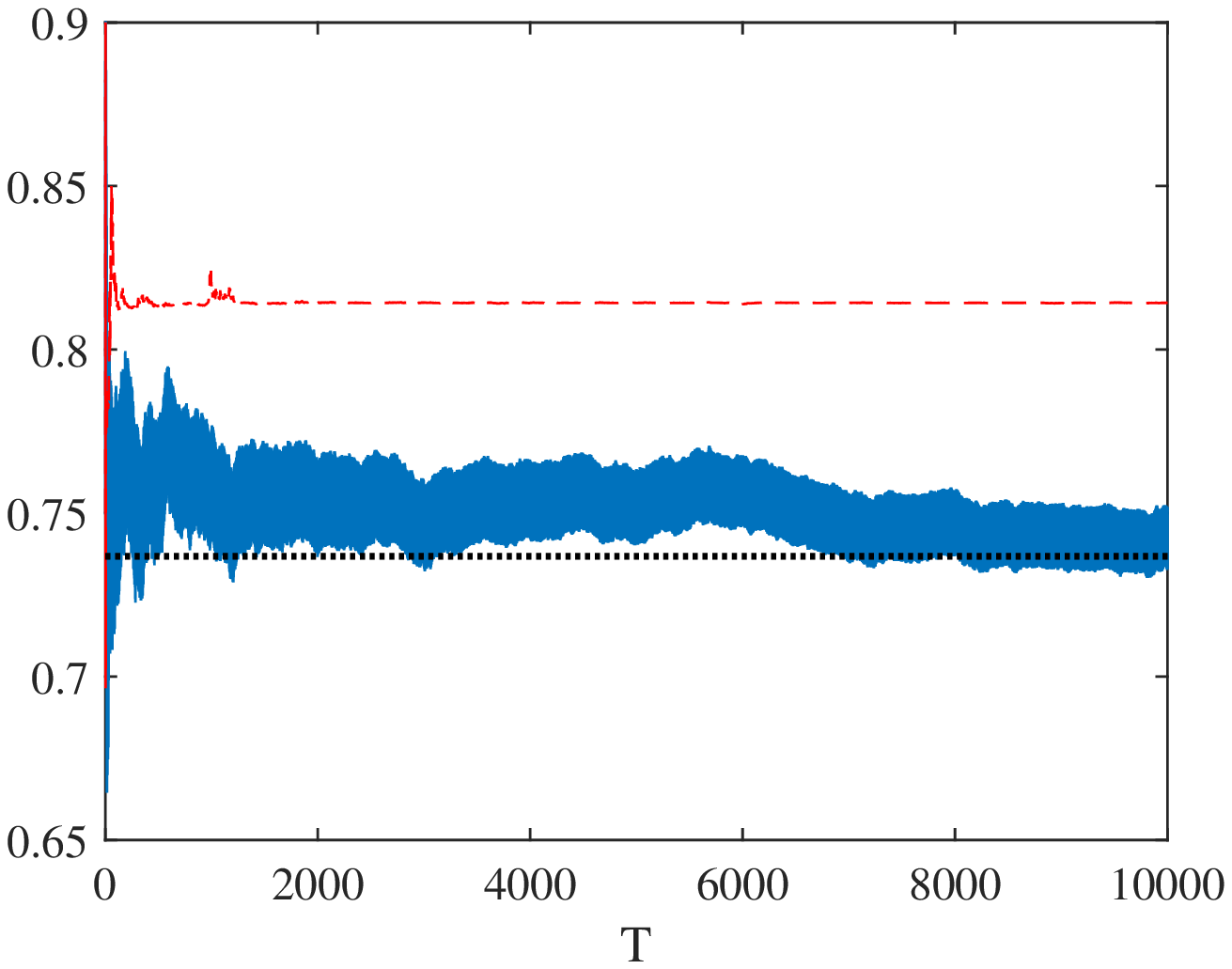}}} 
\subfloat[Mean squared pricing error.]{\label{fig:MSE}\fbox{\includegraphics[width=0.67\columnwidth]{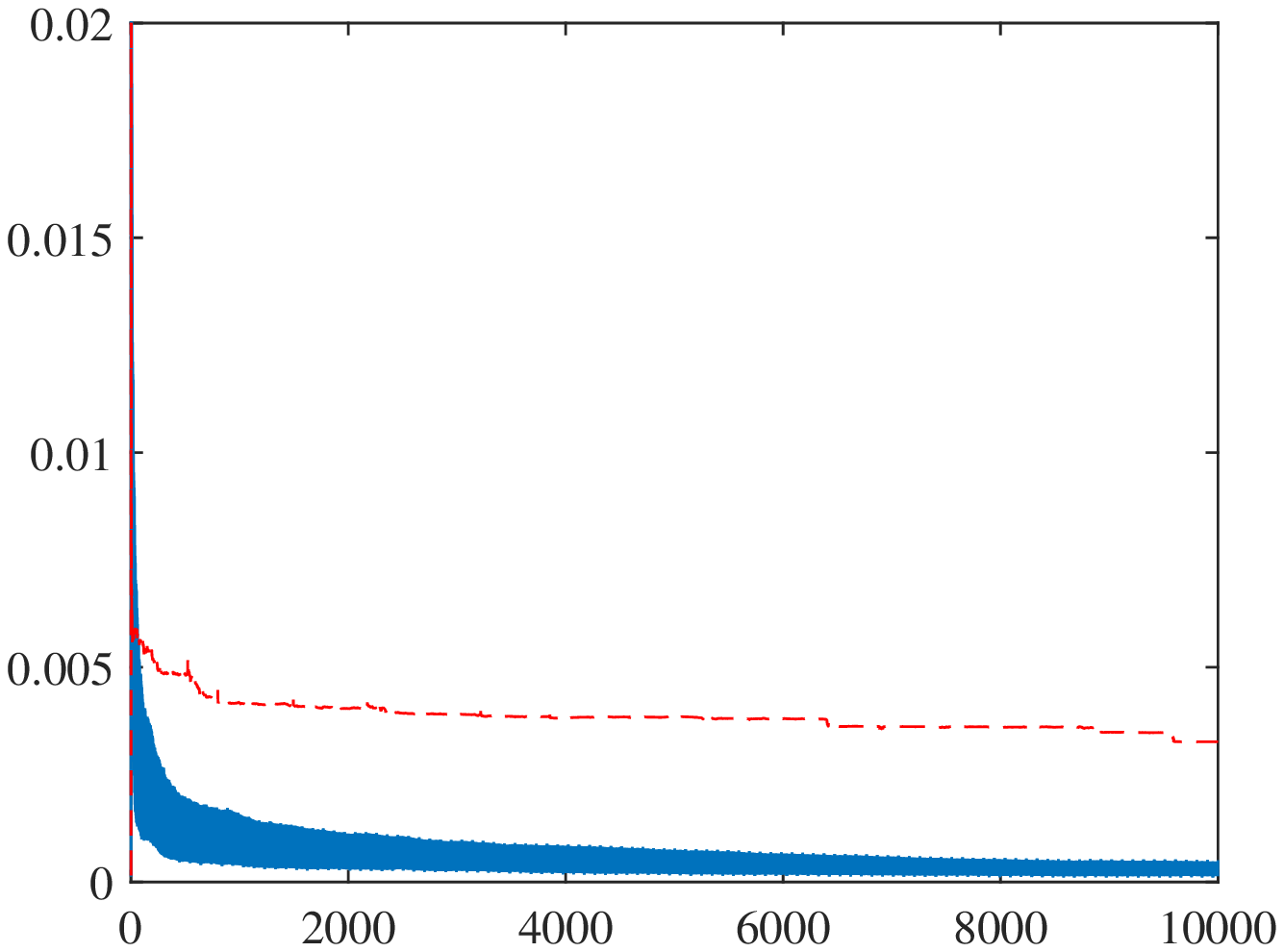}}} 
\subfloat[Regret.]{\label{fig:R}\fbox{\includegraphics[width=0.67\columnwidth]{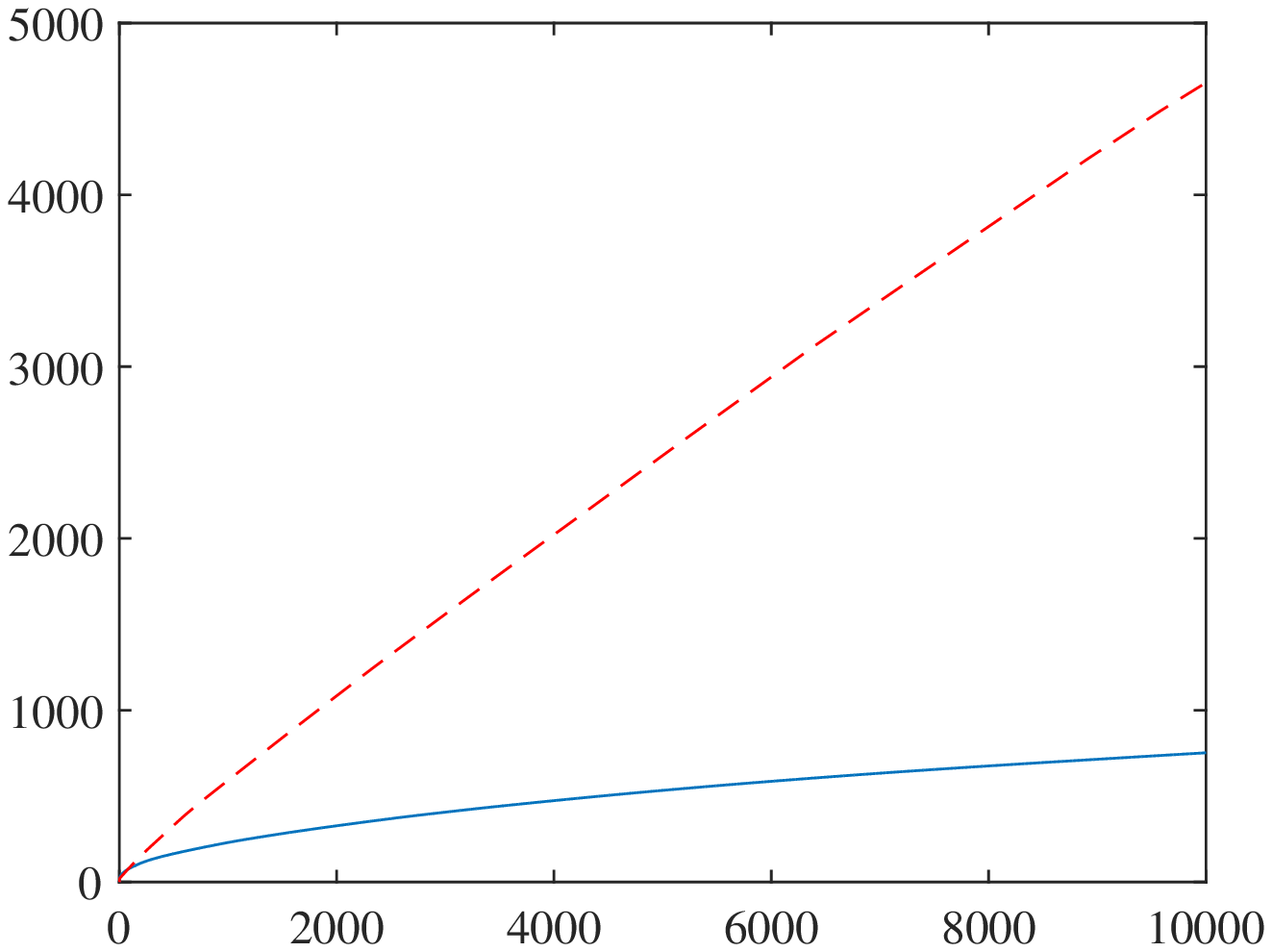}}}
\caption{(a) Sample path of posted prices, (b) mean squared pricing error, and  (c) regret under the \emph{myopic policy} (\red{\hdashrule[0.5ex]{7mm}{1pt}{2.5pt}}), the \emph{perturbed myopic policy} ({\color[rgb]{0,0.4470,0.7410}{\hdashrule[0.5ex]{7mm}{1pt}{}}}), and the \emph{oracle policy} (\hdashrule[0.5ex]{7mm}{1pt}{1pt}).}\label{fig:regret}
\end{figure*}

Because the wholesale price of electricity is fixed over time, the  parameter and quantile estimates represent the  only source of variation in the sequence of prices generated by the myopic policy. Due to the combined structure of the myopic policy and the  least squares estimator,  the value  of each new demand observation rapidly diminishes over time, which, in turn,  manifests in a rapid convergence of the sequence of prices generated under the myopic policy. The resulting lack of exploration in the sequence of  myopic prices  results in incomplete learning, which is seen in Figure \ref{fig:par}.  Namely, the sequence of myopic prices converges to a value, which substantially differs form the oracle optimal price. As a consequence, the myopic policy incurs a $T$-period regret that grows linearly with the horizon $T$, as is observed in Figure \ref{fig:regret}.

On the other hand, the  sequence of perturbations $\{\rho \delta_t\}$ generate enough variation in the sequence of prices generated by the perturbed myopic policy to ensure  consistent model estimation, as is seen in Figures \ref{fig:a} and \ref{fig:b}. This, in turn, results in convergence of the sequence of posted prices to the oracle optimal price. This, combined with the fact that the price offset $\rho \delta_t$ vanishes at a sufficiently fast rate, ensures sublinearity in the growth rate of the corresponding $T$-period regret, as is observed in Figure \ref{fig:regret}.

\section{Conclusion}\label{sec:conclusion}
In this paper, we propose a data-driven approach to pricing demand response with the aim of maximizing the risk-sensitive revenue derived by the electric power utility. The perturbed myopic pricing policy we propose has two key features. First, the unknown demand model parameters are estimated using a least squares estimator. Second, the proposed policy implements a sequence of perturbations to the myopic price sequence to ensure sufficient exploration in the sequence of prices it generates. The price perturbation sequence is designed  to decay at a rate, which is slow enough to ensure complete learning of the underlying demand model, but not so slow  as to preclude a sub-linear growth rate for regret. In particular, the proposed pricing policy is proven to exhibit a $T$-period regret that is no greater than $O(\sqrt{T}\log(T))$. As a direction for future research, it would be interesting to investigate the generalization of the pricing algorithms developed in this paper to  accommodate the treatment of nonlinear and possibly time-varying demand functions.

%%%%%%%%%%%%%%%%%%%%%%%%%%%%%%%%%%%%%%%%%%%%%%%%%%%%%%%%%%%%%%%%%%%%%%%%%%%%%%
%%%%%%%%%%%%%%%%%%%%%%%%%%%%%%%%%%%%%%%%%%%%%%%%%%%%%%%%%%%%%%%%%%%%%%%%%%%%%%
% use bibtex for references
\bibliographystyle{IEEEtran}
\bibliography{references}{\markboth{References}{References}}

\begin{appendices}

\section*{Appendix}

In the following  proofs, we consider a more general form of the perturbation as $\delta_t=\mbox{sgn}(c_t-c_{t-1})\cdot t^{-r},$ where $r$ is allowed to be an arbitrary constant in the interval $[0, 1/2)$. Ultimately, we will prove that a choice of $r=1/4$ minimizes the asymptotic order of the upper bound on regret (ignoring logarithmic factors), which we establish in \eqref{eq:delta r}.

\section{Proof of Lemma \ref{lem:parameters}} \label{app:parameters}
The parameter estimation error derived in Equation \eqref{eq:LSEerror} is given by
\begin{align*}
\theta_t-\theta =  \mathscr{J}_t^{-1} \tilde{\varepsilon}_t,
\end{align*}
where $\tilde{\varepsilon}_t$ is defined as 
\begin{align*}
\tilde{\varepsilon}_t = \sum_{k=1}^t \bmat{p_k \\ 1} \varepsilon_k.
\end{align*}
Using the Cauchy-Schwarz inequality and assuming that $\mathscr{J}_t$ is invertible, the $2$-norm of parameter estimation error is bounded as follows.
\begin{align*}
\| \theta_t-\theta \|^2 &= \| \mathscr{J}_t^{-1} \tilde{\varepsilon}_t \|^2\leq  \| \mathscr{J}_t^{-1/2} \|^2 \| \mathscr{J}_t^{-1/2}\tilde{\varepsilon}_t \|^2.
\end{align*}
Using the definition of matrix norms, we get
\begin{align*}
\| \mathscr{J}_t^{-1/2} \|^2 = \left(\lambda_{\max}(\mathscr{J}_t^{-1/2})\right)^2 = \frac{1}{\lambda_{\min}(\mathscr{J}_t)},
\end{align*}
where the operators $\lambda_{\max}$ and $\lambda_{\min}$ denote the largest and the smallest eigenvalues, respectively. In the following Lemma, we establish a lower bound on the minimum eigenvalue of $\mathscr{J}_t$ in terms of the price perturbations and the wholesale energy price variations.
\begin{lemm}\label{lem:J}
Under the perturbed myopic policy \eqref{policy:unknown}, it holds that
\begin{align}
\lambda_{\min}(\mathscr{J}_t) \geq \frac{1}{1+\overline{p}^2}L_t \quad \text{a.s.}, \label{eq:j l}
\end{align}
where $L_t$ is defined as
\begin{align}
L_t:=\frac{1}{8}\left(\rho^2\lfloor t/2\rfloor^{1-2r}+\sum_{k=1}^{\lfloor t/2\rfloor} \left(c_{2k}-c_{2k-1}\right)^2\right). \label{eq:def l}
\end{align}
\end{lemm}
Using Inequality \eqref{eq:j l}, the mean squared parameter estimation error can be bounded as
\begin{align}
\mathbb{E}\left[\| \theta_t-\theta \|^2\right]&\leq \mathbb{E}\left[\frac{1}{\lambda_{\min}(\mathscr{J}_t)}  \| \mathscr{J}_t^{-1/2}\tilde{\varepsilon}_t \|^2\right]\nonumber\\
&\leq \frac{1+\overline{p}^2}{L_t} \mathbb{E}\left[ \tilde{\varepsilon}_t^\top \mathscr{J}_t^{-1}\tilde{\varepsilon}_t\right]. \label{eq:MSE lmin}
\end{align}
We now establish an upper bound on $\mathbb{E}\left[ \tilde{\varepsilon}_t^\top \mathscr{J}_t^{-1}\tilde{\varepsilon}_t\right]$ by adopting a similar approach as \cite[Lemma 1]{lai1982least}. More specifically, we establish a recursive inequality relating $\mathbb{E}\left[ \tilde{\varepsilon}_t^\top \mathscr{J}_t^{-1}\tilde{\varepsilon}_t\right]$ to $\mathbb{E}\left[ \tilde{\varepsilon}_{t-1}^\top \mathscr{J}_{t-1}^{-1}\tilde{\varepsilon}_{t-1}\right]$. It holds that
\begin{align}
&\mathbb{E}\left[ \tilde{\varepsilon}_t^\top \mathscr{J}_t^{-1}\tilde{\varepsilon}_t\right]\nonumber\\
&\quad=\mathbb{E}\left[ \left(\tilde{\varepsilon}_{t-1}+\bmat{p_t \\ 1} \varepsilon_t\right)^\top \mathscr{J}_t^{-1} \left(\tilde{\varepsilon}_{t-1}+\bmat{p_t \\ 1} \varepsilon_t\right)\right]\nonumber\\
&\quad=\mathbb{E}\left[ \tilde{\varepsilon}_{t-1}^\top \mathscr{J}_t^{-1} \tilde{\varepsilon}_{t-1}\right]+2\mathbb{E}\left[ \tilde{\varepsilon}_{t-1}^\top \mathscr{J}_t^{-1} \bmat{p_t \\ 1} \varepsilon_t\right]\nonumber\\
&\qquad\qquad+\mathbb{E}\left[ \bmat{p_t \\ 1}^\top \mathscr{J}_t^{-1} \bmat{p_t \\ 1} \varepsilon_t^2\right]. \label{eq:lyaponouv}
\end{align}
Using the fact that $\tilde{\varepsilon}_{t-1}$, $p_t$, and $\mathscr{J}_t$ are all measurable according to the $\sigma$-algebra generated by $\varepsilon_1,\ldots,\varepsilon_{t-1}$, and the law of iterated expectations, we get
\begin{align}
&\mathbb{E}\left[ \tilde{\varepsilon}_{t-1}^\top \mathscr{J}_t^{-1} \bmat{p_t \\ 1} \varepsilon_t\right]\nonumber\\
&=\mathbb{E}\left[\mathbb{E}\left[ \tilde{\varepsilon}_{t-1}^\top \mathscr{J}_t^{-1} \bmat{p_t \\ 1} \varepsilon_t\bigg|\varepsilon_1,\ldots,\varepsilon_{t-1}\right]\right]\nonumber\\
&=\mathbb{E}\left[ \tilde{\varepsilon}_{t-1}^\top \mathscr{J}_t^{-1} \bmat{p_t \\ 1} \mathbb{E}\left[\varepsilon_t\big|\varepsilon_1,\ldots,\varepsilon_{t-1}\right]\right]\nonumber\\
&= 0, \label{eq:iterated exp 1}
\end{align}
where the last identity follows from the fact that $\varepsilon_t$ is independent of $\varepsilon_1,\ldots, \varepsilon_{t-1}$ and is zero-mean. Using a similar argument, we get
\begin{align}
\mathbb{E}\left[ \bmat{p_t \\ 1}^\top \mathscr{J}_t^{-1} \bmat{p_t \\ 1} \varepsilon_t^2\right]&= \mathbb{E}\left[ \bmat{p_t \\ 1}^\top \mathscr{J}_t^{-1} \bmat{p_t \\ 1}\right] \mathbb{E}\left[\varepsilon_t^2\right]\nonumber\\
&\leq \mathbb{E}\left[ \bmat{p_t \\ 1}^\top \mathscr{J}_t^{-1} \bmat{p_t \\ 1}\right] \frac{(\overline{\varepsilon}-\underline{\varepsilon})^2}{4},\label{eq:iterated exp 2}
\end{align}
where the last inequality follows from Popoviciu's inequality on variances. By combining Equations \eqref{eq:lyaponouv} and \eqref{eq:iterated exp 1} with Inequality \eqref{eq:iterated exp 2}, we get
\begin{align}
&\mathbb{E}\left[ \tilde{\varepsilon}_t^\top \mathscr{J}_t^{-1}\tilde{\varepsilon}_t\right]\label{eq:mse quad}\\
&\leq \mathbb{E}\left[ \tilde{\varepsilon}_{t-1}^\top \mathscr{J}_t^{-1} \tilde{\varepsilon}_{t-1}\right] + \mathbb{E}\left[ \bmat{p_t \\ 1}^\top \mathscr{J}_t^{-1} \bmat{p_t \\ 1}\right] \frac{(\overline{\varepsilon}-\underline{\varepsilon})^2}{4}.\nonumber
\end{align}
Here, we bound each term in the right hand side of Inequality \eqref{eq:mse quad} separately. For the first term, using the Sherman-Morrison formula, we get
\begin{align*}
\mathscr{J}_t^{-1} &= \left(\mathscr{J}_{t-1}+\bmat{p_t \\ 1}\bmat{p_t \\ 1}^\top\right)^{-1}\\
&=\mathscr{J}_{t-1}^{-1}-\frac{\mathscr{J}_{t-1}^{-1}\bmat{p_t \\ 1}\bmat{p_t \\ 1}^\top\mathscr{J}_{t-1}^{-1}}{1+\bmat{p_t \\ 1}^\top\mathscr{J}_{t-1}^{-1}\bmat{p_t \\ 1}}.
\end{align*} 
Thus,
\begin{align}
&\mathbb{E}\left[ \tilde{\varepsilon}_{t-1}^\top \mathscr{J}_t^{-1} \tilde{\varepsilon}_{t-1}\right]\nonumber\\
&= \mathbb{E}\left[ \tilde{\varepsilon}_{t-1}^\top \mathscr{J}_{t-1}^{-1} \tilde{\varepsilon}_{t-1}\right] - \mathbb{E}\left[\frac{\left(\tilde{\varepsilon}_{t-1}^\top\mathscr{J}_{t-1}^{-1}\bmat{p_t \\ 1}\right)^2}{1+\bmat{p_t \\ 1}^\top\mathscr{J}_{t-1}^{-1}\bmat{p_t \\ 1}}\right]\nonumber\\
&\leq \mathbb{E}\left[ \tilde{\varepsilon}_{t-1}^\top \mathscr{J}_{t-1}^{-1} \tilde{\varepsilon}_{t-1}\right], \label{eq:sherma}
\end{align}
where the equality follows from the fact that the random variable in the second expectation is non-negative almost surely.

For the second term in Inequality \eqref{eq:mse quad}, we have that
\begin{align}
\bmat{p_t \\ 1}^\top \mathscr{J}_t^{-1} \bmat{p_t \\ 1} &= \frac{1}{J_t}\bmat{p_t \\ 1}^\top \bmat{1&-\bar{p}_t\\-\bar{p}_t&(1/t)\sum_{k=1}^t p_k^2} \bmat{p_t \\ 1}\nonumber\\
&=\frac{1}{J_t}\left((p_t-\bar{p}_t)^2+\frac{1}{t}J_t\right)\nonumber\\
&=\frac{J_t-J_{t-1}}{J_t} + \frac{1}{t},\label{eq:quad recursion}
\end{align}
where the last inequality follows from the fact that $J_t-J_{t-1}=(p_t-\bar{p}_t)^2$. Now using Inequalities \eqref{eq:mse quad}, \eqref{eq:sherma}, and \eqref{eq:quad recursion} we get
\begin{align*}
&\mathbb{E}\left[ \tilde{\varepsilon}_t^\top \mathscr{J}_t^{-1}\tilde{\varepsilon}_t\right]\\
&\leq \mathbb{E}\left[ \tilde{\varepsilon}_{t-1}^\top \mathscr{J}_{t-1}^{-1} \tilde{\varepsilon}_{t-1}\right] + \left(\frac{J_t-J_{t-1}}{J_t} + \frac{1}{t}\right) \frac{(\overline{\varepsilon}-\underline{\varepsilon})^2}{4}.
\end{align*}
By summing both sides of the above inequality from $3$ to $t$ we get
\begin{align*}
&\mathbb{E}\left[ \tilde{\varepsilon}_t^\top \mathscr{J}_t^{-1}\tilde{\varepsilon}_t\right]\\
&\leq \mathbb{E}\left[ \tilde{\varepsilon}_{2}^\top \mathscr{J}_{2}^{-1} \tilde{\varepsilon}_{2}\right]+\frac{(\overline{\varepsilon}-\underline{\varepsilon})^2}{4}\mathbb{E}\left[\sum_{k=3}^t \left(\frac{J_k-J_{k-1}}{J_k} + \frac{1}{k}\right) \right].
\end{align*}
It is straightforward to show that
\begin{align*}
\mathbb{E}\left[ \tilde{\varepsilon}_{2}^\top \mathscr{J}_{2}^{-1} \tilde{\varepsilon}_{2}\right]=\mathbb{E}\left[ \varepsilon_{1}^2+\varepsilon_{2}^2\right]\leq \frac{1}{2}(\overline{\varepsilon}-\underline{\varepsilon})^2.
\end{align*}
Note that $\sum_{k=3}^t (1/k) \leq \log(t)$. We also have that
\begin{align*}
\sum_{k=3}^t \frac{J_k-J_{k-1}}{J_k} &= \sum_{k=3}^t \int_{J_{k-1}}^{J_k}\frac{dx}{J_k}\\
&\leq \sum_{k=3}^t \int_{J_{k-1}}^{J_k}\frac{dx}{x}\\
&= \int_{J_{2}}^{J_t}\frac{dx}{x}\\
&\leq \log(J_t)\\
&\leq \log(t\overline{p}^2),
\end{align*}
where the last inequality follows from the fact that $(p_k-\bar{p}_t)^2\leq \overline{p}^2$ almost surely. Finally, we get
\begin{align*}
\mathbb{E}\left[ \tilde{\varepsilon}_t^\top \mathscr{J}_t^{-1}\tilde{\varepsilon}_t\right]&\leq \frac{1}{2}(\overline{\varepsilon}-\underline{\varepsilon})^2\left(1+\log(\bar{p})+\log(t)\right)\\
&\leq \frac{1}{2}(\overline{\varepsilon}-\underline{\varepsilon})^2\left(2+\log(\bar{p})\right) \log(t),
\end{align*}
where the last inequality follows from the fact that $\log(t)\geq 1$ for $t\geq 3$. Finally, by applying the above inequality to the bound on the mean squared parameter estimation error \eqref{eq:MSE lmin}, we get
\begin{align}
\mathbb{E}\left[\| \theta_t-\theta \|^2\right]&\leq \frac{1}{2}(1+\overline{p}^2)(\overline{\varepsilon}-\underline{\varepsilon})^2\left(2+\log(\bar{p})\right)\frac{\log(t)}{L_t}. \label{eq:mse l_t}
\end{align}
To complete the proof, we set $r=1/4$. For this choice of $r$, we have that $L_t\geq \rho^2\sqrt{\lfloor t/2\rfloor}/8\geq \rho^2\sqrt{t}/16$. Setting $\mu_2:=8(1+\overline{p}^2)(\overline{\varepsilon}-\underline{\varepsilon})^2\left(2+\log(\bar{p})\right)$ concludes the proof.

\section{Proof of Theorem \ref{thm:regret-unknown}}\label{app:regret-unknown} 

We introduce an additional assumption on the variation in the sequence of wholesale electricity prices.\footnote{Such assumption will prove useful in facilitating the proof of Theorem  \ref{thm:regret-log}.} Namely, let $\sigma\geq 0$ be  nonnegative constant such that $|c_t-c_{t-1}| \geq \sigma$ for all $t\geq 1$. Ultimately, we will establish the desired result for $\sigma=0$, the setting considered in the statement of the Theorem. 

We begin with the following upper bound on the $T$-period regret.
\begin{align}
&\Delta^\pi(T)=a\sum_{t=1}^{T} \mathbb{E}\left[\left(p_t-p^*_t\right)^2\right]\nonumber\\
&\leq a\sum_{t=1}^{\lfloor \frac{T+1}{2}\rfloor}\mathbb{E}\left[(\widehat{p}_{2t-1}-p^*_{2t-1})^2+(\widehat{p}_{2t-1}-p^*_{2t-1}+\rho\delta_{2t})^2\right]\nonumber\\
&\leq a\sum_{t=1}^{\lfloor \frac{T+1}{2}\rfloor}\left(3\mathbb{E}\left[(\widehat{p}_{2t-1}-p^*_{2t-1})^2\right] + 2\rho^2\delta_{2t}^2 \right)\nonumber\\
&=K_0 + 2a\rho^2\sum_{t=1}^{\lfloor \frac{T+1}{2}\rfloor} (2t)^{-2r} + 3a \sum_{t=1}^{\lfloor \frac{T+1}{2}\rfloor-1}\mathbb{E}\left[(\widehat{p}_{2t+1}-p^*_{2t+1})^2\right], \label{eq:reg sum}
\end{align}
where the second inequality follows from the fact that $x^2+(x+y)^2\leq 3x^2+2y^2$ for any pair of scalars $x, y \in \Rset$. Here, the constant $K_0$ is defined as
\begin{align*}
K_0 = 3a(p_1-p_1^*)^2.
\end{align*}
Recall that $p_1$ is assumed to be a deterministic constant. We now establish upper bounds on each term of the bound \eqref{eq:reg sum} separately. 

\emph{Second term:} \ For all $T\geq 3$, we have that
\begin{align}
\sum_{t=1}^{\lfloor \frac{T+1}{2}\rfloor} (2t)^{-2r} &\leq \int_{0}^{\lfloor \frac{T+1}{2}\rfloor} (2t)^{-2r}dt\nonumber\\
&= \frac{1}{2(1-2r)}\left(2\left\lfloor \frac{T+1}{2}\right\rfloor\right)^{1-2r}\nonumber\\
&\leq \frac{2}{3(1-2r)}T^{1-2r}, \label{eq:T 1-2r}
\end{align}
where the last inequality follows from the fact that $(\frac{T+1}{T})^{1-2r}\leq 4/3$ for all $T\geq 3$ and all $r\in[0,1/2)$.

\emph{Third term:} \ Using the upper bound on the pricing error \eqref{eq:price dev}, we get
\begin{align*}
(\widehat{p}_{2t+1}-p^*_{2t+1})^2\leq 2\kappa_3^2\|\widehat{\theta}_{2t}-\theta \|^2 + 2\kappa_2^2 (F_{2t}^{-1}(\alpha)-F^{-1}(\alpha))^2.
\end{align*}
Then,
\begin{align}
&\sum_{t=1}^{\lfloor \frac{T+1}{2}\rfloor-1}\mathbb{E}\left[(\widehat{p}_{2t+1}-p^*_{2t+1})^2\right]\nonumber\\
&\leq  \sum_{t=1}^{\lfloor \frac{T+1}{2}\rfloor-1}\mathbb{E}\left[2\kappa_3^2\|\widehat{\theta}_{2t}-\theta \|^2 + 2\kappa_2^2 ({F}_{2t}^{-1}(\alpha)-F^{-1}(\alpha))^2\right]\nonumber\\
&\leq \sum_{t=1}^{\lfloor \frac{T+1}{2}\rfloor-1}\left(\kappa_4\frac{\log(2t)}{L_{2t}}+2\kappa_2^2\mathbb{E}\left[({F}_{2t}^{-1}(\alpha)-F^{-1}(\alpha))^2\right]\right), \label{eq:reg MSE}
\end{align}
where the last inequality follows from the upper bound \eqref{eq:mse l_t} on the mean squared parameter estimation error and $\kappa_4:=(1+\overline{p}^2)(\overline{\varepsilon}-\underline{\varepsilon})^2\left(2+\log(\bar{p})\right)\kappa_3^2$. Using the fact that for a continuous nonnegative random variable $X$, it holds that $\mathbb{E}[X]=\int_0^\infty \mathbb{P}\{X\geq x\}dx$, we get
\begin{align}
\mathbb{E}\left[({F}_{2t}^{-1}(\alpha)\hspace{-0.075em}-\hspace{-0.075em}F^{-1}(\alpha))^2\right]\hspace{-0.2em}&=\hspace{-0.3em}\int_0^\infty \hspace{-0.75em} \mathbb{P}\{({F}_{2t}^{-1}(\alpha)\hspace{-0.075em}-\hspace{-0.075em}F^{-1}(\alpha))^2 \hspace{-0.1em}\geq\hspace{-0.1em} \gamma\}d\gamma\nonumber\\
&\leq \int_0^\infty 2\exp(-\mu_1 \gamma (2t)) d\gamma\nonumber\\
&=\frac{1}{\mu_1 t},\label{eq:mse quantile}
\end{align} 
where the inequality follows from the bound \eqref{eq:dvoretzky}. By combining Inequalities \eqref{eq:reg sum}, \eqref{eq:T 1-2r}, \eqref{eq:reg MSE}, and \eqref{eq:mse quantile}, we get
\begin{align}
&\Delta^\pi(T)\nonumber\\
&\leq K_0 + \frac{4a}{3(1-2r)}\rho^2 T^{1-2r} + 3a \sum_{t=1}^{\lfloor \frac{T+1}{2}\rfloor-1}\left(\kappa_4\frac{\log(2t)}{L_{2t}}+\frac{2\kappa_2^2}{\mu_1 t}\right)\nonumber\\
&\leq  K_0 +  \frac{4a}{3(1-2r)}\rho^2 T^{1-2r}+24a\kappa_4 \sum_{t=1}^{\lfloor \frac{T+1}{2}\rfloor-1}\frac{\log(2t)}{\rho^2 t^{1-2r}+\sigma^2t} \nonumber  \\
& \qquad+ \frac{6a\kappa_2^2}{\mu_1}(1+\log(T)),\label{eq:delta r}
\end{align}
where the last inequality follows from the definition of $L_{2t}$ in Equation \eqref{eq:def l} and the assumption that $|c_t-c_{t-1}|\geq \sigma$ for all $t$. For $\sigma =0$, it is straightforward to show that a choice of $r=1/4$  minimizes the asymptotic order of the upper bound \eqref{eq:delta r} with respect to the horizon $T$ up to multiplicative logarithmic factors. Setting $r=1/4$ and $\sigma=0$  yields
\begin{align}
\Delta^\pi(T)&\leq  K_1  + K_2\log(T)+ K_3 \rho^2\sqrt{T} + \frac{K_4}{\rho^2}  \sum_{t=1}^{\lfloor \frac{T+1}{2}\rfloor-1} \frac{\log(2t)}{\sqrt{t}},\label{eq:delta r=1/4}
\end{align}
where $K_1:=K_0 + K_2$, $K_2 := {6a\kappa_2^2}/{\mu_1}$, $K_3 :=8a/3$, and $K_4 := 24a\kappa_4$. It holds that
\begin{align*}
\sum_{t=1}^{\lfloor \frac{T+1}{2}\rfloor-1} \frac{\log(2t)}{\sqrt{t}}&\leq \log(2) + \sum_{t=2}^{\lfloor \frac{T+1}{2}\rfloor-1} \frac{\log(2t)}{\sqrt{t}}\\
&\leq \log(2) +  \int_1^{T/2} \frac{\log(2t)}{\sqrt{t}}dt\\
&\leq  \log(2) +  2 \sqrt{\frac{T}{2}}\log\left(T\right).
\end{align*}
Finally, we define the nonnegative constants $C_0$, $C_1$, and $C_2$ as follows to conclude the proof.
\begin{align}
&C_0:=  K_1 + \frac{K_4\log(2)}{\rho^2}\label{eq:C_0}\\
&C_1:= \frac{\sqrt{2}K_4}{\rho^2}+K_3\rho^2 \label{eq:C_1}\\
&C_2:= K_2. \label{eq:C_2}
\end{align}

\section{Proof of Theorem \ref{thm:regret-log}}\label{app:regret-log}
Inequality  \eqref{eq:delta r} is a valid upper bound on the $T$-period regret incurred by perturbed myopic policy, under the assumption that $\sigma>0$. By setting $\rho=0$, the upper bound simplifies to
\begin{align*}
\Delta^\pi(T)&\leq K_0 +24a\kappa_4 \sum_{t=1}^{\lfloor \frac{T+1}{2}\rfloor-1}\frac{\log(2t)}{\sigma^2t} + \frac{6a\kappa_2^2}{\mu_1}(1+\log(T)),
\end{align*}
It holds that
\begin{align*}
\sum_{t=1}^{T/2} \frac{\log(2t)}{t}\leq \log(2)+\int_1^{T/2} \frac{\log(2t)}{t}dt\leq \log(2)+\log^2(T).
\end{align*}
We define the nonnegative constants $M_0$, $M_1$, and $M_2$   as follows to conclude the proof.
\begin{align}
M_0&:= K_0 + M_1  \label{eq:M_0}\\
M_1&:=\frac{6a\kappa_2^2}{\mu_1} \label{eq:M_1}\\
M_2&:= 24a\kappa_4. \label{eq:M_2}
\end{align}

\section{Proof of Lemma \ref{lem:J}}\label{app:J}
It is straightforward to show that the characteristic polynomial of $\mathscr{J}_{t}$ is given by
\begin{align*}
\lambda^2-\lambda\left(t+\sum_{k=1}^t p_k^2\right) + t J_t=0.
\end{align*}
Then, 
\begin{align*}
\lambda_{\max}(\mathscr{J}_{t})+\lambda_{\min}(\mathscr{J}_{t}) &= t+\sum_{k=1}^t p_k^2,\\
\lambda_{\max}(\mathscr{J}_{t})\lambda_{\min}(\mathscr{J}_{t}) &=t J_t.
\end{align*}
From the first identity it follows that 
\begin{align*}
\lambda_{\max}(\mathscr{J}_{t})\leq t+\sum_{k=1}^t p_k^2\leq t(1+\overline{p}^2).
\end{align*}
Thus, we get
\begin{align*}
\lambda_{\min}(\mathscr{J}_{t})=\frac{t J_t}{\lambda_{\max}(\mathscr{J}_{t})}\geq \frac{J_t}{1+\overline{p}^2}.
\end{align*}
We now bound the random process $\{J_t\}$ from below by a deterministic sequence. Fix $t$. A direct substitution of the perturbed myopic policy yields
\begin{align*}
J_t&\geq \sum_{k=1}^{\lfloor t/2\rfloor} \Bigg\{\left(\widehat{p}_{2k-1}-\bar{p}_{t}\right)^2\\
&\hspace{4em}+\left(\widehat{p}_{2k-1}-\bar{p}_{t}+\frac{1}{2}\left(c_{2k}-c_{2k-1}\right)+\rho\delta_{2k}\right)^2\Bigg\}.
\end{align*}
The above inequality can be further relaxed to eliminate its explicit dependency on the (random) price process. Namely, it is straightforward to show that
\begin{align}
J_t&\geq \frac{1}{2}\sum_{k=1}^{\lfloor t/2\rfloor}\frac{\rho^2}{(2k)^{2r}}+\frac{1}{8}\sum_{k=1}^{\lfloor t/2\rfloor} \left(c_{2k}-c_{2k-1}\right)^2\label{eq:int J}.
\end{align}
One can further relax  inequality  \eqref{eq:int J} by using the facts that \[\sum_{k=1}^t \frac{1}{k^{2r}}\geq \int_1^{t+1} \frac{1}{x^{2r}}dx=\frac{(t+1)^{1-2r}-1}{1-2r},\] and \[(t+1)^{1-2r}-1  \geq  t^{1-2r}\left(1-\frac{1}{2^{1-2r}}\right).\] 
It follows that
\begin{align}
J_t&\geq \frac{\rho^2}{2^{1+2r}}\frac{\lfloor t/2\rfloor^{1-2r}}{1-2r}\left(1-\frac{1}{2^{1-2r}}\right)+\frac{1}{8}\sum_{k=1}^{\lfloor t/2\rfloor} \left(c_{2k}-c_{2k-1}\right)^2\nonumber\\
&\geq L_t, \label{eq:bound J}
\end{align}
where $L_t$ is defined as \[L_t:=\frac{1}{8}\left(\rho^2\lfloor t/2\rfloor^{1-2r}+\sum_{k=1}^{\lfloor t/2\rfloor} \left(c_{2k}-c_{2k-1}\right)^2\right).\]

\end{appendices}

\end{document}